%% file: main.tex
\documentclass[preprint]{elsarticle}

\usepackage{comment}
\usepackage{graphicx}
\usepackage{booktabs}
\usepackage{xspace}
\usepackage{xcolor}
\usepackage{tabularx}
\usepackage{latexsym,amssymb,amsmath,amsthm}
\usepackage{float}

\DeclareMathAlphabet{\altmathcal}{OMS}{cmsy}{m}{n}

\usepackage{newtxtext,newtxmath}
\usepackage[scaled=.85]{newtxtt}

\usepackage[defblank]{paralist}
\usepackage{enumerate}
\usepackage{hyperref}

\usepackage{listings}

\usepackage{tikz}
\usetikzlibrary{matrix,positioning,calc}

\input{defs}
\input{comments}

\usepackage[strings]{underscore}

\newtheorem{example}{Example}

\journal{Data \& Knowledge Engineering}

\begin{document}

\begin{frontmatter}
  \title{Conceptually-grounded Mapping Patterns for Virtual Knowledge Graphs 
}
%
\author[inst1,inst2]{Diego Calvanese}
\author[inst3]{Avigdor Gal}
\author[inst1]{Davide Lanti\corref{correspondingauthor}}
\cortext[correspondingauthor]{Corresponding author}
\ead{lanti@inf.unibz.it}
\author[inst1]{Marco Montali}
\author[inst1]{Alessandro Mosca}
\author[inst4]{Roee Shraga}

\address[inst1]{Free-University of Bozen-Bolzano, Bolzano, Italy}
\address[inst2]{Ume\aa\ University, Sweden}
\address[inst3]{Technion -- Israel Institute of Technology, Haifa, Israel}
\address[inst4]{Khoury College of Computer Science, Northeastern University, Boston, Massachusetts}

\begin{abstract}
  Virtual Knowledge Graphs (VKG) constitute one of the most promising paradigms
  for integrating and accessing legacy data sources. A critical bottleneck in
  the integration process involves the definition, validation, and maintenance
  of mapping assertions that link data sources to a domain ontology.
  To support the management of mappings throughout their entire lifecycle, we
  identify a comprehensive catalog of sophisticated mapping patterns that
  emerge when linking databases to ontologies. To do so, we build on
  well-established methodologies and patterns studied in data management, data
  analysis, and conceptual modeling. These are extended and refined
  through the analysis of concrete VKG benchmarks and real-world
  use cases, and considering the inherent impedance mismatch between data
  sources and ontologies.  We validate our catalog on the
  considered VKG scenarios, showing that it covers the vast majority of
  mappings present therein.
\end{abstract}

  \begin{keyword}
    Virtual Knowledge Graphs \sep Ontology-based Data Access \sep
    Mapping patterns \sep Data Integration.
  \end{keyword}
\end{frontmatter}

\input{1-introduction}
\input{2-preliminaries}
\input{3-patterns}
\input{3.2-mapping-schema.tex}
\input{3.3-mapping-data.tex}
\input{3.4-variations.tex}
\input{3.5-automatic-discovery.tex}
\input{5-pattern-usage}
\input{6-scenarios}
\input{7-related-long}
\input{8-conclusions}

\mypar{Acknowledgements}
This research has been partially supported by the EU H2020 project INODE (grant
agreement No~863410), by the Italian PRIN project HOPE,
and by the Free University of Bozen-Bolzano through the MP4OBDA
project. Diego Calvanese also acknowledges the support of the Wallenberg AI,
Autonomous Systems and Software Program (WASP) funded by the Knut and Alice
Wallenberg Foundation and Avigdor Gal the support of the Benjamin and Florence
Free Chair.

\bibliographystyle{elsarticle-num}
\bibliography{string-tiny,w3c,krdb,local-bib}

\end{document}

%% file: defs.tex
\makeatletter
\newcommand\footnoteurl@[1]{\footnote{\url@{#1}}}
\DeclareRobustCommand{\footnoteurl}{\hyper@normalise\footnoteurl@}
\makeatother

\definecolor{bostonuniversityred}{rgb}{0.8, 0.0, 0.0}
\definecolor{awesome}{rgb}{1.0, 0.13, 0.32}
\definecolor{darkorange}{rgb}{1.00,0.55,0.00}

\newcommand{\at}{\operatorname{at}}

\newcommand{\nullable}{\textsl{opt}}

\newcommand{\er}{E-R\xspace}
\newcommand{\uml}{UML\xspace}
\newcommand{\orm}{ORM~2\xspace}

\newcommand{\dm}{W3C-DM\xspace}

\newcommand{\ontop}{Ontop\xspace}

\newcommand{\NC}{\textbf{NC}\xspace}
\newcommand{\NP}{\textbf{NP}\xspace}
\newcommand{\NI}{\textbf{NI}\xspace}
\newcommand{\ND}{\textbf{ND}\xspace}

\renewcommand{\vec}[1]{\mathbf{#1}}

\newcommand{\tup}[1]{\langle #1\rangle}
\newcommand{\ex}[1]{\textsf{#1}}
\newcommand{\pat}[1]{\textbf{#1}}

\newcommand{\bsbm}{BSBM\xspace}
\newcommand{\npd}{NPD\xspace}
\newcommand{\uobm}{UOBM\xspace}
\newcommand{\odh}{ODH\xspace}
\newcommand{\stod}{ST-OD\xspace}
\newcommand{\cordis}{Cordis\xspace}
\newcommand{\nn}{--}

\newcommand{\partition}{\textsl{part}}

\newcommand{\iri}{\mathfrak{t}}
\newcommand{\pk}[1]{\underline{#1}}

\newcommand{\key}[2][]{\textsf{key}_{#1}(#2)}

\newcommand{\fd}[3][]{\textsf{fd}(#1{:}\ #2 \rightarrow #3)}

\newcommand{\ISA}{\sqsubseteq}

\newcommand{\rdf}{\textsc{RDF}\xspace}
\newcommand{\shacl}{\textsc{SHACL}\xspace}
\newcommand{\owlql}{\textsc{OWL\,2\,QL}\xspace}
\newcommand{\owl}{\textsc{OWL\,2}\xspace}
\newcommand{\owlel}{\textsc{OWL\,2\,EL}\xspace}
\newcommand{\owlrl}{\textsc{OWL\,2\,RL}\xspace}

\newcommand{\dlliteR}{\textit{DL-Lite}$_{\altmathcal{R}}$\xspace}

\newcommand{\vkg}{VKG\xspace}
\newcommand{\vkgs}{VKGs\xspace}

\newcommand{\sparql}{\textsc{SPARQL}\xspace}
\newcommand{\sql}{\textsc{SQL}\xspace}

\newcommand{\setsym}[1]{\textsf{\textbf{#1}}}
\newcommand{\sA}{\setsym{A}}
\newcommand{\sB}{\setsym{B}}

\newcommand{\sK}{\setsym{K}}

\newcommand{\sU}{\setsym{U}}

\newcommand{\A}{\mathcal{A}}
\newcommand{\C}{\mathcal{C}}
\newcommand{\D}{\mathcal{D}}
\newcommand{\I}{\mathcal{I}}

\newcommand{\M}{\mathcal{M}}

\renewcommand{\O}{\mathcal{O}}
\renewcommand{\S}{\mathcal{S}}
\newcommand{\T}{\mathcal{T}}

\newcommand{\setB}[2]{\left\{ #1 \mid #2\right\}}

\newlength{\erw}
\newcommand{\erscale}{0.6}
\newlength{\schemaw}
\newlength{\uoffset}
\newlength{\uuoffset}
\newlength{\doffset}
\newlength{\ddoffset}

\tikzstyle{dbpicture}=[remember picture, baseline, node distance=0mm,
  thick, font=\footnotesize]

\newcommand{\mypar}[1]{\smallskip\noindent\textbf{#1}.~}


%% file: comments.tex
\newif\ifdraft
\drafttrue

\ifdraft
\newcounter{todocntr}
\newenvironment{todoenv}
{\addtocounter{todocntr}{1}\par\bigskip\noindent
  \color{blue}\fbox{\thetodocntr}\hspace*{\fill}~\bf
  \begin{minipage}[t]{0.95\linewidth}}
  {\end{minipage}\hspace*{\fill}\bigskip}
\newcommand{\todo}[1]{\begin{todoenv}{#1}\end{todoenv}}
\newcommand{\td}[1]{\noindent{\color{red}\framebox[\linewidth]{\parbox{0.95\linewidth}{\bf #1}}}}
\newcommand{\boxcomment}[3]{~\\{\color{#2}{\fbox{\parbox{\linewidth}{{#1}:~#3}}}~\\}}
\makeatletter
\if@twocolumn
\usepackage{marginnote}
\newcommand{\nb}[2]{%
  \makebox[0cm][r]{\textcolor{#2}{\bf!}}%
  \marginnote[{\scriptsize\textcolor{#2}{#1}}]%
  {{\scriptsize\textcolor{#2}{#1}}}%
}
\newcommand{\inlinecomment}[3]{
  \marginnote[\textcolor{#2}{#1}]{\textcolor{#2}{#1}}\textcolor{#2}{#3}%
}
\else
\newcommand{\nb}[2]{\makebox[0cm][r]{\textcolor{#2}{\bf!}}%
  \marginpar[\parbox{\marginparwidth}{\scriptsize\textcolor{#2}{\raggedleft #1}}]%
  {\parbox{\marginparwidth}{\scriptsize\textcolor{#2}{\raggedright #1}}}}
\newcommand{\inlinecomment}[3]{
  \marginpar[\textcolor{#2}{#1}]{\textcolor{#2}{#1}}\textcolor{#2}{#3}%
}
\fi
\makeatother
\else
\newcommand{\todo}[1]{}
\newcommand{\td}[1]{}
\newcommand{\nb}[2]{}
\newcommand{\inlinecomment}[3]{}
\newcommand{\boxcomment}[3]{}
\fi



%% file: 1-introduction.tex
\section{Introduction}
\label{sec:introduction}


Data integration and access to legacy data sources using end user-oriented languages are increasingly challenging contemporary organizations. In the whole spectrum of data integration and access solutions, the approach based on \emph{Virtual Knowledge Graphs (\vkgs)} is gaining momentum~\cite{2021Hogan}, especially when the underlying data sources to be integrated come in the form of relational databases (DBs)~\cite{XDCC19}. \vkgs replace the rigid structure of tables with the flexibility of a graph that incorporates domain knowledge and is kept virtual, eliminating the need of making a copy of the data as in a typical ETL-based (Extract, Transform, Load) approach, thus avoiding duplication of data and guaranteeing the freshness of the information being accessed. A \vkg specification consists of three main components:
\begin{inparaenum}[\it (i)]
\item \emph{data sources} (in the context of this paper, constituted by relational DBs) where the actual data are stored;
\item a domain \emph{ontology}, capturing the relevant concepts, relations, and constraints of the domain of interest;
\item a set of \emph{mappings} linking the data sources to the ontology.
\end{inparaenum}
One of the most critical bottlenecks towards the adoption of the \vkg approach, especially in complex, enterprise scenarios, is the definition and management of mappings.

  Mappings play a central role in a variety of data management tasks, within both the Semantic Web and the DB communities, and come in different forms. In \emph{schema matching}, for example, mappings (typically referred to as ``matches'') aim at expressing correspondences between atomic, constitutive elements of two different relational schemata, such as attributes and relation names~\cite{rahm2001survey}. In this context, very sophisticated (semi-)automatic techniques are being developed to bootstrap this simple type of mappings, without prior knowledge on the two schemata~\cite{do2002coma,chen2018biggorilla,shragaadnev}. A similar setting arises in the context of \emph{ontology matching} (also referred to as \emph{ontology alignment}), where the atomic elements to be put into correspondence are concepts and properties~\cite{EUZENAT2007a}. Just like with schema matching, a huge body of applied research has led to effective (semi-)automatic techniques for establishing mappings~\cite{ivanova2017alignment,kolyvakis2018deepalignment}.
In \emph{data exchange}, instead, more complex mapping specifications (like the well-known formalism of TGDs~\cite{AbHV95,DLLR07,Kola05}) are needed to express how data extracted from a source DB schema should be used to populate a target DB schema \cite{Lenz02}. Due to the complex nature of these mappings, research in this field has been mainly foundational, with few notable exceptions \cite{FHHM*09,CKQT18}.

The \vkg approach appears to be the one that poses the most advanced challenges when it comes to mapping specification, debugging, and maintenance. On the one hand, \vkg mappings are inherently more sophisticated than those used in schema and ontology matching. On the other hand, while they appear to resemble those typically used in data exchange, they need to overcome the abstraction mismatch between the relational schema of the underlying data storage, and the target ontology; consequently, they are required to explicitly handle how (tuples of) data values extracted from the DB lead to the creation of corresponding objects in the ontology.

It is not surprising, then, that management of \vkg mappings throughout
their entire lifecycle is currently a labor-intensive and mostly manual
effort, which requires highly-skilled professionals~\cite{spanos2012bringing} that, at once:
\begin{inparaenum}[\itshape (i)]
\item  have in-depth knowledge of the domain of discourse and how it can be represented using structural conceptual models (such as UML class diagrams) and ontologies;
\item  possess the ability to understand and query the logical and physical structure of the DB; and
\item  master languages, methodologies, and technologies for representing the ontology and the mappings using standard frameworks from Semantic Web (such as the OWL 2 profiles and R2RML).
\end{inparaenum}
Even in the presence of all these skills, writing mappings is demanding and
poses a number of challenges related to semantics, correctness, and
performance. More concretely, no comprehensive approach currently exists to
support ontology engineers and knowledge scientists~\cite{FletcherGrothSequeda2020} in the creation of \vkg mappings, exploiting all the involved information artifacts to their full potential: the \textbf{relational schema} with its constraints and the extensional information stored in the DB, the \textbf{ontology axioms}, and a \textbf{conceptual model} that lies, explicitly or implicitly, at the basis of the relational schema.

Bootstrapping techniques~\cite{JKZH*15,KHSB*17} have been developed to relieve the ontology engineer from the ``blank paper syndrome''. However, they are typically adopted in scenarios where neither the ontology nor the mappings are initially available, and various assumptions are posed over the schema of the DB (e.g., in terms of normalization). Hence, they essentially bootstrap at once the ontology as a ``query-preserving''~\cite{SequedaArenasDM} mirrored image of the DB, and the corresponding one-to-one mappings. These approaches typically work at the level of DB schemata, ignoring the data, and therefore they might fail in those cases were the DB schema is either poorly structured or the applicable constraints are not fully specified.

Most of research so-far has been focused on bootstrapping, and a common trait of all these works is that they gloss over or altogether ignore the actual conceptual model underlying the given database instance, arguing that such conceptual model is in many cases not available to the bootstrapper. This choice, however, leads to ambiguities and arbitrary choices: the same relational schema might, in fact, correspond to several different conceptual representations. We here take a completely different approach: we single out mapping patterns by fully accounting for the intended conceptual representation and the database schema corresponding to such representation. This allows us to define each pattern in a non-ambiguous way, and to precisely specify the constraints to be imported into the \vkg\footnote{Modulo the expressive power of the ontology language being considered.}.

\begin{figure}[t]
  \label{f:db-conc-onto}
  \centering
  \includegraphics[width=.9\textwidth]{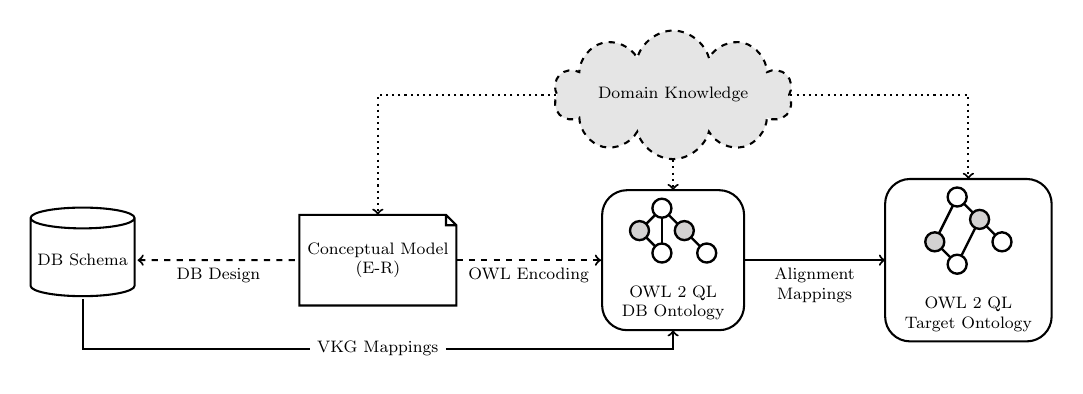}
  \caption{The database and the ontology both stem from common domain knowledge.}
\end{figure}

The justification for our approach is depicted in Figure~\ref{f:db-conc-onto}, and
foresees a scenario where both the ontology and the DB schema are
derived from a conceptual analysis of the domain of interest. The resulting
knowledge may stay implicit, or lead
to an explicit representation in the form of a structured conceptual model~\cite{CKMSA18},
usually represented using well-established notations such as \uml, \orm, or
\er. On the one hand, this conceptual model provides the basis for creating a
corresponding domain ontology through a series of 
transformation steps, where these steps should ideally preserve the semantics of the original model, modulo the expressivity of the considered ontology language~\cite{CaLN98,BeCD05,HaMo10}.
On the other hand, it can trigger the design process that
finally leads to the deployment of an actual DB. This is done via a
series of restructuring and adaptation steps, considering a number of aspects
that go beyond pure conceptualization, such as query load, performance, volume,
and the abstraction gap that exists between the conceptual
and logical/physical layers. It is precisely the reconciliation of these two
transformation chains (resp., from the conceptual model to the ontology, and
from the conceptual model to the DB) that is reflected in the \vkg
mappings.

From this key observation, we derive a catalog of \emph{mapping patterns} that emerge when linking DBs to ontologies.
To do so, we build on well-established methodologies and patterns studied in
data management (such as W3C Direct Mapping -- \dm \cite{W3Crec-RDB-Direct-Mapping} -- and its extensions), data analysis (such as algorithms for discovering dependencies), and conceptual modeling (such as well-known \emph{lossless} transformations from E-R diagrams to DB schemata~\cite{Codd71,HaMo10Chapter11,EmTh11Chapter5}).
These are suitably extended and refined, by considering the inherent impedance mismatch between
data sources and ontologies, which requires to handle the creation of objects
starting from DB values.

The idea of mapping patterns is not new, and was first introduced in~\cite{sequeda2012}, later refined in~\cite{sequeda-book}. However, there are substantial differences between the patterns discussed in this line of works and the patterns we introduce here. Specifically, those patterns are usually informally specified and quite permissive, not grounding the DB instance to any particular conceptual representation. This allows the KG practitioner to map KGs that are potentially very different from the intended DB conceptualization. On the contrary, each of our patterns explicitly and non-ambiguously specifies the link between the conceptualization and the DB instance, which is the one arising from applying well-known and \emph{semantics-preserving} transformations studied in the area of DB design. Our choice is motivated by the observation that the conceptual structures ``encoded'' in the DB instance are not arbitrary, but derive from the design phase of the DB and reflect the actual domain knowledge, and from the fact that the expressive power of the ontology language commonly adopted for \vkgs is comparable to that of E-R diagrams.

This foundational grounding, which clearly distinguishes our work from related literature, paves the way towards a variety of \vkg design scenarios, depending on which information artifacts are available, and which ones must be produced. For example, our patterns could be used to validate existing mappings, to generate mappings and an ontology if only the DB is available, or even also as a basis for reconstructing the conceptual model when it is left implicit or inaccessible.

Another major contribution of this work, which to the best of our knowledge distinguishes it from all the related literature, is an evaluation of our approach. Concretely, we analyze six concrete \vkg scenarios and benchmarks, and report on the coverage of mappings appearing therein in terms of our patterns, as well as on how many times the same pattern recurs. This also gives an interesting
indication on which patterns are more pervasively used in practice.

As a final remark, the patterns we introduce here were developed while satisfying the restrictions imposed by the VKG setting (e.g., at the level of expressiveness of the ontology language). However, they can be directly applied in all those contexts where such restrictions are satisfied, regardless of whether the knowledge graph is virtual.

The remainder of this paper is structured as follows: Section~\ref{sec:preliminaries} introduces the notation and basic notions on \vkgs; Section~\ref{sec:patterns} contains our catalog of mapping patterns, the main contribution of this work; Section~\ref{sec:discussion} discusses possible applications of the catalog, depending on the information artifacts that are available to the \vkg designer; Section~\ref{sec:scenarios} presents an evaluation of the catalog over six different scenarios of different complexities; Section~\ref{sec:related} discusses related work, and Section~\ref{sec:conclusions} concludes the paper.




%% file: 2-preliminaries.tex
\section{Preliminaries}
\label{sec:preliminaries}

In this work, we use the \textbf{bold} font to denote tuples, e.g., $\vec{x}$,
$\vec{y}$, are tuples. When convenient and non-ambiguous, we treat tuples as sets and
use set operators on them.

We assume that the reader is familiar with standard notions and languages from the
relational databases world, such as SQL or \er diagrams. Other readers might want
to refer to the abundant literature on the subject, with~\cite{AbHV95} as an excellent primer.

We rely on the \vkg framework as previously introduced in \cite{PLCD*08}.
A {\vkg specification} is a triple $(\T,\M,\S)$
where $\T$ is an \emph{ontology TBox}, $\M$ is a set of
\emph{\vkg mappings} (or, simply, \emph{mappings}), and $\S$ is a DB schema. We next introduce these elements and their semantics.

The schema $\S$ is a pair $(\Sigma, \Gamma)$. $\Sigma$ is the \emph{signature} of $\S$, that is,
a set of \emph{table schemata} of the form $T(A_1, \ldots, A_n)$, where $T$ is
a \emph{table name} and $A_1, \ldots, A_n$ are \emph{attributes}, each associated to a \sql \emph{datatype}. $\Gamma$ is a set of database constraints. In this work, $\Gamma$ consists
of \emph{key} and \emph{foreign key} constraints, as well as \emph{inclusion dependencies}.
In this work, foreign keys are inclusion dependencies where the \emph{referred} attributes form a (not necessarily \emph{primary}) key\footnote{Although some systems, like \texttt{MySQL}, do not impose particular restrictions over the referred attributes, the common assumption in the literature is that the referred attributes form a \emph{primary key} (e.g., see~\cite{silberschatz-7}).}.
A \emph{database instance} $\D$ for $\S$ is
a first-order interpretation mapping each table name to a relation over the interpretation domain, and that \emph{satisfies}\footnote{For details, refer to classic literature such as ~\cite{AbHV95}.} the constraints in $\Gamma$.

The ontology $\T$ is formulated in \owlql~\cite{W3Crec-OWL2-Profiles}, but for conciseness we here use its logical underpinning, \dlliteR~\cite{CDLLR07}, slightly enriched to handle datatypes.

\paragraph{Syntax}
We fix four enumerable, pairwise-disjoint
sets: \NI of \emph{individuals}, \NC of \emph{class names}, \NP of \emph{object
 property names}, and \ND of \emph{data property names}. An \owlql
\emph{TBox} $\T$ is a finite set of \emph{axioms} of the following form:
{\small
\[
  \boxed{
    \begin{array}{l@{\hspace{1cm}}l}
      B \ISA C & q \ISA r \\
      \rho(d) \ISA f & d \ISA v
    \end{array}
  }
\]
}
where $B, C$ are \emph{classes}, $q$, $r$ are \emph{object properties}, $d$ is a \emph{data property}, $f$ is a \emph{datatype expression}, $\rho(d)$ is a \emph{data property range expression}, and $v$ is a \emph{data property expression}. Elements above are defined according to the following grammar, where $A \in \NC$,  $d \in \ND$, $p \in \NP$, $\delta(d)$ is a \emph{data property domain expression}, and $T_1, \ldots, T_n$ are the RDF datatypes\footnote{The RDF and OWL Recommendations use the simple types from XML Schema (\url{https://www.w3.org/TR/xmlschema-2/})}:
{\small
\[
  \boxed{
    \begin{array}{l@{\qquad\qquad}l}
      B ~\rightarrow~ A ~\mid~ \exists r ~\mid~ \delta(d) & C ~\rightarrow~ \top_C ~\mid~ B ~\mid~ \lnot B \\
      q ~\rightarrow~ p ~\mid~ p^- & r ~\rightarrow~ q ~\mid~ \lnot q\\
      f ~\rightarrow~ \top_D ~\mid~ T_1 ~\mid~ \cdots ~\mid~ T_n & v ~\rightarrow~ d ~\mid~ \lnot d
    \end{array}
  }
\]
}
In the rules above, $\top_C$ and $\top_D$ denote the ``top'' concepts for concepts and data values (called \emph{literals} in the RDF terminology), respectively.

There are a few differences between the (\dlliteR-like) language introduced here and \owlql. Specifically, \owlql also allows one to express special features of binary relations, such as reflexivity or transitivity. Since these additional constructs do not affect our patterns, they are not part of the ontology language considered here.

An \owlql \emph{ABox} $\A$ is a finite set of \emph{assertions} of the form $A(a)$, $p(a,b)$, or $d(a,\ell)$, where $A\in\NC$, $p\in\NP$, $d\in\ND$, $a$ and $b$ individuals in $\NI$, and $\ell$ a literal value.  We call the pair $\O=\tup{\T,\A}$ an \owlql \emph{ontology}.

\paragraph{Semantics}

Similarly to first-order logic, the semantics of \owlql ontologies is given through Tarski-style
\emph{interpretations} $\I = (\Delta_O^\I, \Delta_V^\I, \cdot^\I)$, where $\Delta_O^\I$ is a non-empty \emph{domain of objects}, $\Delta_v^\I$ is a non-empty \emph{domain of values}, and $\cdot^\I$ is an interpretation function defined according to the inductive definition of Table~\ref{t:owlql-sem}.
\begin{table}[t] \centering
  \scriptsize
  \begin{tabular}{l l l l}
    \toprule
    Construct & Syntax Element & Example & Semantics \\
    \midrule
    Top class & $\top_C$ & & $\Delta_O^\I$  \\[.3em]
    Top domain & $\top_V$ & & $\Delta_V^\I$  \\[.3em]
    Concept name & $A \in \NC$ & \sf{Person} & $A^\I \subseteq \Delta_O^\I$ \\[.3em]
    Object property name & $p \in \NP$ & \sf{hasSpouse} & $p^\I \subseteq \Delta_O^\I \times \Delta_O^\I$ \\[.3em]
    Inverse object property & $p^-$ & \sf{hasSpouse}$^-$ & $\setB{(y,x)}{(x,y) \in p^\I}$ \\[.3em]
    Data property name & $d \in \ND$ & \sf{hasName} & $d^\I \subseteq \Delta_O^\I \times \Delta_V^\I$ \\[.3em]
    Datatype & $T_i$ & \sf{xsd:int} & $T_i^\I \subseteq \Delta_V^\I$ \\[.3em]
    Existential restriction & $\exists r$ & $\exists \sf{hasSpouse}$ & $\setB{x \in \Delta_O^\I}{\exists y \in \Delta_O^\I : (x, y) \in r^\I}$ \\[.3em]
    Data property domain & $\delta(d)$ & $\delta(\sf{hasName})$ & $\setB{x \in \Delta_V^\I}{\exists v \in \Delta_V^\I : (x, v) \in d^\I}$ \\[.3em]
    Data property range & $\rho(d)$ & $\rho(\sf{salary})$ & $\setB{v \in \Delta_V^\I}{\exists o \in \Delta_O^\I : (o, v) \in d^\I}$ \\[.3em]
    Concept negation & $\lnot A$ & $\lnot \sf{Human}$ & $\Delta_O^I \setminus A^\I$ \\[.3em]
    Object property negation & $\lnot r$ & $\lnot \sf{hasSpouse}$ & $\Delta_O^I \times \Delta_O^\I \setminus r^\I$ \\[.3em]
    Data property negation & $\lnot d$ & $\lnot \sf{hasName}$ & $\Delta_O^I \times \Delta_V^\I \setminus d^\I$ \\[.3em]
    Individual & $a \in \NI$ & \sf{george} & $a^\I \in \Delta_O^\I$ \\[.3em]
    Literal & $\ell \in \ND$ & \sf{``george''} & $\ell^\I \in \Delta_V^\I$ \\
    \bottomrule
  \end{tabular}
  \caption{\label{t:owlql-sem}Semantics for \owlql constructs.}
\end{table}

Let $\I$ be an interpretation. \emph{$\I$ satisfies an inclusion axiom $\phi \sqsubseteq \psi$}, denoted as $\I \models \phi \sqsubseteq \psi$, if $\phi^\I \subseteq \psi^\I$. \emph{$\I$ satisfies a TBox $\T$}, denoted as $\I \models \T$, if it satisfies all axioms in $\T$. \emph{$\I$ satisfies a class assertion $C(a)$}, denoted as $\I \models C(a)$, if $a^\I \in C^\I$. \emph{$\I$ satisfies an object property assertion $R(a,b)$ (resp., a data property assertion $d(a, \ell)$)}, denoted as $\I \models r(a,b)$ (resp., $\I \models d(a, \ell)$), iff $(a^\I,b^\I) \in r^\I$ (resp., $(a^\I,\ell^\I) \in d^\I$). \emph{$\I$ satisfies an ABox $\A$}, denoted as $\I \models \A$, if it satisfies all assertions in $\A$. Finally, \emph{$\I$ satisfies an ontology $\O = (\T, \A)$}, denoted as $\I \models \O$, if $\I \models \T$ and $\I \models \A$.

\paragraph{Mappings}

In the \vkg literature, mappings specify how to populate classes and properties of the ontology with
individuals and values constructed from the data in the underlying
DB. In other words, mappings provide the ABox that, together with a given TBox, realizes an ontology. In \vkgs, the adopted language for mappings in real-world systems is
R2RML~\cite{W3Crec-R2RML}, but for conciseness we use here a more convenient
abstract notation inspired by the literature~\cite{PLCD*08}: a \emph{mapping} $m$ is a pair of the form
{\small
\[
  \boxed{
    \begin{array}{l@{\qquad\qquad}l}
      s: Q(\vec{x}) &
      t: \vec{L}(\vec{\iri}(\vec{x}))
    \end{array}
  }
\]
}
where $Q(\vec{x})$ is a SQL query over the DB schema $\S$, called \emph{source query}, and $\vec{L}(\vec{\iri}(\vec{x}))$ is a list of \emph{target atoms} of the form
\begin{inparablank}
\item $C(\iri_1(\vec{x_1}))$,
\item $p(\iri_1(\vec{x_1}), \iri_2(\vec{x_2}))$, or
\item $d(\iri_1(\vec{x_1}), \iri_2(\vec{x_2}))$,
\end{inparablank}
where $C \in \NC$, $p \in \NP$, $d \in \ND$, and $\iri_1(\vec{x_1})$ and $\iri_2(\vec{x_2})$ are terms that we call \emph{templates}. In this work we express source queries using the notation of
\emph{relational algebra} omitting answer variables under the assumption that
they coincide with the variables used in the target atoms. Intuitively, a
template $\iri(\vec{x})$ in the target atom of a mapping corresponds to an
R2RML \emph{string template}\footnote{\url{https://www.w3.org/TR/r2rml/\#dfn-string-template}}, and is used to generate object \emph{IRIs} (Internationalized Resource Identifiers) or (RDF) \emph{literals}, starting from DB values retrieved by the source query in that mapping. Note that, R2RML constants can be rendered in our syntax by using templates of zero arity, and that data values (e.g., using the {rr:column} predicate) can be simulated by introducing a dedicated special template symbol.

Given a set $\M$ of mappings, and a database instance $\D$, the \emph{virtual ABox $\A_{\M(\D)}$ exposed by $\D$ through $\M$} is the set of ABox assertions:
\[\setB{\vec{L}(\at(\vec{\iri}(\vec{x}), (\vec{x} \mapsto \vec{o})))}{(\vec{x} \mapsto \vec{o}) \in Q(\vec{x})^\D, (s: Q(\vec{x}), t: \vec{L}(\vec{\iri}(\vec{x}))) \in \M}\]
where $(\vec{x} \mapsto \vec{o})$ is a \emph{solution mapping} belonging to the evaluation $Q(\vec{x})^\D$ of the source query $Q(\vec{x})$ over the DB instance $\D$, and $\vec{L}(\at(\vec{\iri}(\vec{x}), (\vec{x} \mapsto \vec{o})))$ is a set of ABox assertions deriving from the \emph{application of the template} $\vec{\iri}(\vec{x})$ over the solution mapping $(\vec{x} \mapsto \vec{o})$. Such template application is usually defined by replacing $\vec{x}$ with $\vec{o}$ in $\vec{t}(\vec{x})$, through \emph{string concatenation} operations.

In the following, we provide an example of mapping and virtual ABox derived through it. For the examples, we will use the concrete syntax adopted by the \ontop \vkg system~\cite{CCKK*17}, in which the source query is expressed in \sql and each target atom is expressed as an \emph{\rdf triple pattern with templates}. The answer variables of the source query occurring in the target atoms are distinguished by enclosing them in curly brackets \verb|{|\,$\cdots$\verb|}|.
The following is an example mapping expressed in such syntax:
\begin{lstlisting}[basicstyle=\ttfamily\footnotesize]
@prefix ex: <http://www.example.com/> .
source    SELECT ssn FROM person
target    ex:person/{ssn} a ex:Person .
\end{lstlisting}
The first line is a prefix declaration, used to abbreviate URIs. For instance, \texttt{ex:Person} is an abbreviation for the URI \texttt{http://www.example.com/Person}.
The effect of such mapping, when applied to a DB instance $\D$ for
$\Sigma$, is to populate the class \texttt{ex:Person} with IRIs constructed by
replacing the answer variable \texttt{ssn} occurring in the
target atom with the corresponding assigments for that variable in the solution mappings
to the source query evaluated over $\D$. For instance, if the \sql query
in the source retrieves the solution mappings (\texttt{ssn} $\mapsto$ \texttt{000-00-0000}) and (\texttt{ssn} $\mapsto$ \texttt{000-00-0001}), then the mapping above produces
the following RDF graph (expressed in the Turtle~\cite{W3Crec-RDF-1.1-Turtle} syntax):
\begin{lstlisting}
@prefix ex: <http://www.example.com/> .
ex:person/000-00-0000 a ex:Person .
ex:person/000-00-0001 a ex:Person .
\end{lstlisting}
stating that individuals \texttt{ex:person/000-00-0000} and \texttt{ex:person/000-00-0001} are both instances of class \ex{ex:Person}.

Given a \vkg specification $(\T, \M, \S)$ and a database instance $\D$ of $\S$, the ontology $\O = (\T, \A_{\M(\D)})$ is called \emph{Virtual Knowledge Graph of $(\T, \M, \S)$ through $\D$}. The name ``virtual'' in the name derives from the fact that the virtual ABox $\A_{\M(\D)}$ in a \vkg setting is not materialized and stored somewhere. Query answering in \vkg, in fact, is carried out through \emph{query rewriting and query unfolding techniques}~\cite{PLCD*08,CCKK*17}: \sparql queries get translated on-the-fly into equivalent SQL queries, that are directly evaluated against the DB, transparently to the end-user.

\subsection{R2RML Mappings vs \vkg Mappings}
\label{s:r2rml-vs-vkg}

In this work, we adopt an abstract syntax for mappings inspired by the well-established scientific literature on \vkgs, e.g.~\cite{PLCD*08}. On the other hand, R2RML was inspired by the works on RDB2RDF systems\footnote{\url{https://www.w3.org/2001/sw/rdb2rdf/}}, and does not enforce some of the conventions which are common (and, sometimes, required) in a \vkg setting. Usually these low-level differences are never addressed explicitly in the scientific literature, but in our experience of maintainers of the \ontop~\cite{CCKK*17} VKG system we have witnessed that these can be a major source of confusion for VKG practitioners. Since this work is also targeting potential users of a VKG system that will have to write concrete R2RML mappings, we here explicitly clarify what these conventions are.

We recognize two main differences between R2RML Mappings and VKG Mappings: the first is that R2RML Mappings support what we here call \emph{intensional mappings}, and the second one is that R2RML also allows for templates that are not \emph{injective}.

\paragraph{Intensional Mappings}

R2RML allows for mappings at the \emph{intensional level}, i.e., for mappings defining TBox axioms rather than ABox assertions, like the following one (expressed in the \ontop~\cite{CCKK*17} concrete mapping syntax):

\begin{lstlisting}
@prefix ex: <http://www.example.com/> .
@prefix rdfs: <http://www.w3.org/2000/01/rdf-schema#> .
source    SELECT star_id, star_type FROM star
target    ex:star/{star_type} rdfs:subClassOf :Star .
\end{lstlisting}

Mappings like the one above go beyond those considered in the scientific literature for \vkgs, and are often a source of confusion among \vkg practitioners: they are perfectly valid R2RML mappings, hence virtually all R2RML-based \vkg systems accepts them, however the semantics of query answering in the presence of such mappings is system-dependent, often leading to ``surprising'' results which are due to the well-studied computational complexity boundaries of the \vkg approach~\cite{PLCD*08,ACKZ09}. Here, we ignore such kind of mappings, and focus instead on traditional \vkg mappings.

\paragraph{Injective Templates} In classic \vkg literature~\cite{PLCD*08}, templates are always \emph{injective}, that is, the application of different templates (i.e., templates differing in either function symbol or arity) can never generate the same \emph{ground term}\footnote{In DL literature, logical first-order terms are used in place of URIs or literals from the RDF world.}. This detail is \emph{usually}\footnote{An exception is the notion of ``OBDA-complete mapping'' introduced in~\cite{phd-thesis}.} never made explicit, because it is a trivial consequence of the fact that, in those settings, equality between terms is realized through syntactic \emph{unification}. R2RML deals with URIs, not with logical terms. Hence, it is inherently more flexible, and different URI templates do not guarantee that two different URIs will be generated. As an example, consider the following pair of mappings:

\begin{lstlisting}
@prefix : <http://www.example.com/> .
@prefix rdfs: <http://www.w3.org/2000/01/rdf-schema#> .
source    SELECT star_id, star_type FROM star
target    :star-{star_id} a :Star .

@prefix : <http://www.example.com/> .
@prefix rdfs: <http://www.w3.org/2000/01/rdf-schema#> .
source    SELECT star_id1, star_id2 FROM binary_systems
target    :star-{star_id1}-{star_id2} a :BinaryStarSystem .
\end{lstlisting}

Assume a database instance such that the SQL query in the first mapping retrieves (\texttt{star\_id} $\mapsto$ \texttt{001-100}), and the SQL query in the second mapping retrieves (\texttt{star\_id1} $\mapsto$ \texttt{001}, \texttt{star\_id2} $\mapsto$ \texttt{100}). Then, these solutions together with the mapping assertions above would contruct the following RDF graph:

\begin{lstlisting}
  @prefix : <http://www.example.com/> .
  :star-001-100 a :Star .
  :star-001-100 a :BinaryStarSystem .
\end{lstlisting}

That is, object \texttt{:star-001-100} is both recognized as a single star and and as a binary system composed of two stars. Note that this does not correspond to the information that was contained in the original DB instance: it might be the result of a careful analysis of the domain (hence, the original DB was either incomplete, or not in 1st Normal Form), or it could be just a mistake induced by the fact of not having used injective templates. If the latter, to guarantee injectivity it would have sufficed to use a \emph{safe URI separator} in place of the dash ``-'' in the target of the second mapping. As a matter of fact, the R2RML recommendation states that designers \emph{should}\footnote{\url{https://www.w3.org/TR/r2rml/\#dfn-template-valued-term-map}} use safe URI separators: if this recommendation is observed, then the injectivity condition is trivially satisfied.

Contrarily to the issue of mappings at the intensional level discussed above, the \vkg specification can naturally be extended to the scenario of non-injective templates. However, adopting non-injective templates usually worsens the performance of query reformulation, since joins over individuals built out of different templates are not anymore guaranteed to be empty. Non-injective templates can become useful in particular applications where the URIs to be produced for objects identifiers should conform to some global vocabulary~\cite{CLFMX21}. In this work we adopt the standard assumption of injective templates, as we are interested in preserving the semantics of the original data instance.



%% file: 3-patterns.tex
\section{Mapping Patterns}
\label{sec:patterns}

We now enter into the core contribution of this paper, namely the catalog of \emph{mapping patterns}.
In our vision, (ontology) mapping patterns can be used to unravel the high-level conceptualization behind the database design, and exploit this conceptualization to better link the content of the database to a domain ontology, according to the vision displayed in Figure~\ref{f:db-conc-onto}. 
To justify our formalization of patterns, we make the following two fundamental observations:
\begin{compactenum}[\itshape(i)]
\item a conceptual model may have more than one admissible relational representations, according to the applied methodology, as well as to considerations about efficiency, performance optimization, and space consumption on the final information system;
\item given the logical schema of a relational database, regardless of its normal form, multiple conceptual models can provide (admissible) alternative representations of its domain.
\end{compactenum}
By~\emph{(i)} and~\emph{(ii)}, and differently from most other approaches in the literature, our patterns explicitly specify the target conceptual model, in order to disambiguate among the various admissible conceptualizations for the same database schema.
%

Our patterns are tailored towards the VKG settings, however they can be directly applied also in those, possibly non-virtualized, contexts requiring a lightweight ontology language. The only strict requirements are that the target ontology language is not more expressive than OWL~2~QL, and that the mapping language conforms to the restrictions set in place in Section~\ref{sec:preliminaries}. Studying patterns under different assumptions, e.g., more expressive ontology languages, goes beyond the scope of this work.

In its basic form, a mapping pattern is a quadruple $(\C,\S,\M,\O)$ where $\C$ is a conceptual model, $\S$ is a database schema, $\M$ is a set of mappings, and $\O$ is an (\owlql) ontology.
In such pattern, the pair $(\C, \S)$ puts into correspondence a conceptual representation to one of its (many) admissible (i.e., formally sound~\cite{Hull86,MiIR94}) database schemata, like those prescribed by \emph{well-established database modeling methodologies}. 
The pair $(\M, \O)$, instead, is formed by the \emph{DB ontology} $\O$, which is the \owlql encoding\footnote{Modulo the expressivity of the \owlql language.} of the conceptual model $\C$, and the set $\M$ of mappings, providing the link between $\S$ and $\O$. The term ``DB ontology'' refers to an ontology whose concepts and properties reflect the constructs of the conceptual model, mirroring the structure of the relational database, as displayed in Figure~\ref{f:db-conc-onto}.

As pointed out in Section~\ref{sec:introduction}, we do not fix which of these information artifacts are given, and which are produced as output, but we simply describe how they relate to each other, on a per-pattern basis. Inputs and outputs will depend on the specific application scenario, as we will discuss in Section~\ref{sec:scenarios}.

Some of the more advanced patterns will have a more complex structure, where pairs of conceptual models and/or pairs of database schemata are respectively used in place of $\C$ and $\S$ (e.g., Pattern \pat{SRR} or \pat{SHa} that we will introduce in Section~\ref{sec:mappings} fall under this category). These patterns prescribe specific \emph{transformations} to be applied on an \emph{input} conceptual (resp., DB) schema, in order to obtain an \emph{output} conceptual (resp., DB) schema. These output artifacts make explicit the presence of specific structures that are revealed through the application of the pattern itself. We will see that these structures can in turn enable further applications of patterns.


\subsection{Patterns Organization and Presentation Conventions}

\paragraph{Patterns Organization}
We organize patterns in two major groups: \emph{schema-driven patterns}, shaped
by the structure of the DB schema and its explicit constraints, and
\emph{data-driven patterns}, which in addition consider constraints emerging
from specific configurations of the data in the DB.
Observe that, for each schema-driven pattern, we actually identify a corresponding
data-driven version in which the constraints over the schema are not
explicitly specified, but \emph{hold in the data}. We denote such pattern as its
schema-driven counterpart, but with a leading ``D'' in place of ``S'' (e.g., \pat{DE} is the data-driven version of Pattern \pat{SE} shown in Table~\ref{tab:schema-driven-patterns-1}).


\paragraph{Presentation Conventions}
We show the fragment of the conceptual model that is affected by the pattern
in E-R notation (adopting the original notation by Chen~\cite{Chen76}) -- but any structural
conceptual modeling language, such as \uml or \orm~\cite{UML17,HaMo10}, would work as well. To
compactly represent sets of attributes, we use a small diamond in place of the
small circle used for single attributes in Chen notation. For cardinality constraints we follow
the ``\emph{look-here}'' convention, that is, the cardinality constraint for a \emph{role} is placed
next to the entity participating in that role.
In the DB schema, we
use $T(\pk{\sK},\sA)$ to denote a \emph{table} with name $T$, \emph{primary
 key} consisting of the attributes $\sK$, and additional attributes
$\sA$. Given a set $\sU$ of attributes in $T$, we denote by $\key[T]{\sU}$ the
fact that $\sU$ form a \emph{key} for $T$. Referential integrity constraints
(like, e.g., foreign keys) are denoted with edges, pointing from the
referencing attribute(s) to the referenced one(s). For conciseness, we denote
sets of the form $\{o \mid \mathit{condition}\}$ as
$\{o\}_{\mathit{condition}}$. In order to express datatypes for data properties, we introduce two auxiliary functions: a function $\tau$ that, given a DB attribute $A$, returns the DB datatype of $A$, and a function $\mu$ that associates, to each DB datatype, a corresponding RDF datatype. For the definition of $\mu$, we re-use the \textit{Natural Mapping}\footnoteurl{https://www.w3.org/TR/r2rml/#natural-mapping} correspondence provided by the R2RML recommendation, and displayed in Table~\ref{t:natural-mapping}. As a final note, following the E-R-diagrams convention, we assume a default $(1,1)$ cardinality on attributes. For such a reason, in the DB schema we assume all attributes to be \emph{not nullable} by default (using the \sql convention, declared as ``\texttt{NOT NULL}''). If attribute $A$ is instead optional, we denote this fact in the DB schema through the notation $\nullable(A)$. Such notation extends in the natural way to the case where $A$ is a set of attributes.

\begin{table}
  \centering
  \scriptsize
  \begin{tabular}[t]{ll}
    \toprule
    \textsc{SQL Datatype} & \textsc{RDF Datatype} \\
    \midrule
    String-like & xsd:string \\
    BINARY, BINARY VARYING, BINARY LARGE OBJECT & xsd:hexBinary \\
    NUMERIC, DECIMAL & xsd:decimal \\
    SMALLINT, INTEGER, BIGINT & xsd:integer \\
    FLOAT, REAL, DOUBLE PRECISION & xsd:double \\
    BOOLEAN &   xsd:boolean \\
    DATE & xsd:date \\
    TIME & xsd:time \\
    TIMESTAMP & xsd:dateTime \\
    \bottomrule
  \end{tabular}
    \caption{\label{t:natural-mapping}R2RML Natural Mapping}
\end{table}


%% file: 3.2-mapping-schema.tex
\subsection{Schema-driven Patterns}\label{sec:mappings}

\input{tables/schema-driven-1.tex}
\input{tables/schema-driven-2.tex}

Next we comment on schema-driven patterns, shown in Tables~\ref{tab:schema-driven-patterns-1} and~\ref{tab:schema-driven-patterns-2}. For each pattern we provide an example and references to the relevant related literature. We point out the reference to the literature is \emph{only with respect to the pattern being considered}. Hence, patterns that have been proposed or that can be identified in the related literature, but that do not find a correspondence with our patterns, are out of the scope of this section. These and other considerations will instead be discussed in Section~\ref{sec:related}.

\mypar{Schema Entity (SE)} This fundamental pattern describes the correspondence between an entity with a primary identifier and attributes to a class and data properties in the ontology. The entity is expressed in the DB schema through a single table $T_E$ with primary key $\sK$ and other attributes $\sA$, as it is the norm in sound DB design practices.
The mappings column explains how $T_E$ is mapped into a
corresponding class $C_E$.  The primary key of $T_E$ is employed to construct
the objects that are instances of $C_E$, using a template $\iri_E$ specific for
that entity. Each \emph{relevant} attribute of $T_E$ is mapped to a data property of
$C_E$, with suitable domain and range axioms. A mandatory participation constraint is added to
the each data property corresponding to a mandatory attribute.
\\
\textsl{Example}: A client registry table containing \emph{social security numbers} (SSNs) of clients, together
with their name as an additional attribute, is mapped to a \ex{Client} class
using the SSN to construct its objects. In addition, the SSN and name are
mapped to two corresponding data properties.

\begin{center}
\resizebox{.9\textwidth}{!}{%
  \begin{tabular}{c@{\hspace{2pt}}c}
    \toprule
  \textsc{Conceptual Model} & \textsc{DB schema}\\
  \midrule
  \begin{minipage}{\erw}
    \centerline{\includegraphics[scale=1]{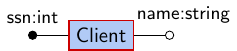}}
  \end{minipage}
  &
    \ex{client(\underline{ssn:int}, name:string)}
  \\[1em]
  \midrule
  \textsc{Mappings} & \textsc{Ontology} \\
  \midrule
  \begin{tabular}{l@{~}l}
      \ex{s{:}} & \ex{SELECT * FROM client} \\
      \ex{t{:}} & \ex{:client/\{ssn\} a :Client} . \\
                & \ex{:client/\{ssn\} :client\#ssn \{ssn\}} .\\
                & \ex{:client/\{ssn\} :client\#name \{name\}} .\\
    \end{tabular}
  &
    \begin{tabular}{ll}
      :Client a owl:Class . \\
      :client\#ssn a owl:DatatypeProperty . & 
      :client\#name a owl:DatatypeProperty . \\
      :client\#ssn rdfs:domain :Client . &
      :client\#ssn rdfs:range xsd:integer . \\
      :client\#name rdfs:domain :Client . &
      :client\#name rdfs:range xsd:string . \\
      :Client rdfs:subClassOf _:r1 . &
      _:r1 a owl:Restriction . \\
      _:r1 owl:onProperty :client\#ssn . &
      _:r1 owl:someValuesFrom rdfs:Literal . \\
      :Client rdfs:subClassOf _:r2 . &
      _:r2 a owl:Restriction . \\
      _:r2 owl:onProperty :client\#name . &
      _:r2 owl:someValuesFrom rdfs:Literal .
    \end{tabular} \\
  \bottomrule
\end{tabular}
}
\end{center}

~\\
\noindent\textsl{References}: This is the most basic pattern, and variants or portions of it are widespread in other approaches.
\begin{compactitem}
\item \dm states that each table with a primary key must be transformed into a class similarly to \pat{SE}. This approach does not fix the conceptual model, nor it considers ontology axioms.
\item The extension of \dm with \owl discussed in~\cite{SequedaArenasDM} includes domain and range axioms for properties, but not mandatory participation of data properties. It includes, however, \owl \emph{key} axioms (\texttt{owl:hasKey}), which we ignore here because they fall outside of the \owlql profile. As for \dm, also this approach does not fix the conceptual model underlying the DB schema.
\item Pattern \pat{SE} resembles to the combination of cases \emph{non-binary relation} and \emph{data attribute} discussed in BootOX~\cite{JKZH*15}, however there are also substantial differences: the catalog of BootOX does not fix the underlying conceptual representation, and as a result ambiguities can arise. For instance, \pat{SE} adds a mandatory participation constraint for a data property only if such property is mapped to a mandatory attribute, whereas in BootOX the choice of having or not each constraint is left up to the user.
\item The algorithm of MIRROR~\cite{mirror} applies a variant of this pattern where the datatype information is encoded into the R2RML mappings, and not added as ontology axioms. Mandatory participation for properties relative to \emph{not nullable} attributes cannot be encoded through this approach.
\item Among the mapping patterns from~\cite{sequeda-book}, our \pat{SE} resembles the union of \emph{Direct Concept} and \emph{Direct Concept Attribute}. Ontology axioms are not included.
\end{compactitem}

\mypar{Schema Relationship (SR)} %
This pattern describes the correspondence between a binary relationship without attributes to an \owlql object property, for the case where such relationship is represented in the DB as a separate (usually, ``Many-to-Many'' table). This pattern considers three tables $T_R$, $T_E$, and $T_F$, in which the set of
columns in $T_R$ is partitioned into two parts $\sK_{RE}$ and $\sK_{RF}$
that are foreign keys to $T_E$ and $T_F$, respectively. The identifier of $T_R$
depends on the role cardinalities in the E-R model. The pattern captures how $T_R$ is mapped to
an object property $p_R$, using the two partitions $\sK_{RE}$ and $\sK_{RF}$
to construct respectively the subject and the object of the triples
in $p_R$. The templates $\iri_{C_E}$ and $\iri_{C_F}$ must be those used for building
instances of classes $C_E$ and $C_F$, respectively.
\\
\textsl{Example}: An additional table in the client registry stores the
addresses of each client, and has a foreign key to a table with locations.
The former table is mapped to an \ex{address} object property, for which the ontology asserts that the domain is the class \ex{Person} and the range an additional class \ex{Location}, which corresponds to the latter table.
\begin{center}
\noindent \resizebox{.9\textwidth}{!}{%
\begin{tabular}{c@{\hspace{2pt}}c}
  \toprule
  \textsc{Conceptual Model} & \textsc{DB schema}\\
  \multicolumn{2}{c}{(Datatypes Omitted)} \\
  \midrule
  \begin{minipage}{\erw}
    \centerline{\includegraphics[scale=1]{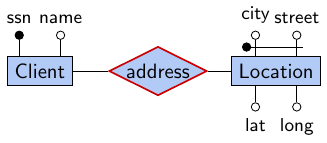}}
  \end{minipage}
  &
    \begin{tikzpicture}[dbpicture]
      \node (e)
      {\normalsize\ex{client}(\pk{\ex{s\tikz\coordinate(kE);sn}},\ex{name})};
      \node[right=of e] (f)
      {\normalsize\ex{location}(\ex{\pk{city, \tikz\coordinate(kF);street}, lat, long})} ;
      \node[below=of e.south west,anchor=north west] (r)
      {\normalsize\ex{address}(\pk{\ex{cli\tikz\coordinate(kRE);ent, ccity,\tikz\coordinate(kRF); cstreet}})};
      \draw[->] ($(kRE)+(0,\uoffset)$) -- ($(kRE)+(0,\uuoffset)$) -| ($(kE)-(0,\doffset)$);
      \draw[->] ($(kRF)+(0,\uoffset)$) |- ($(kRF)+(0,\uuoffset)$) -|
      ($(kF)-(0,\doffset)$);
    \end{tikzpicture}
  \\[2em]
  \midrule
  \textsc{Mappings} & \textsc{Ontology} \\
  \midrule
  \begin{tabular}{l@{~}l}
      \ex{s{:}} & \ex{SELECT * FROM address} \\
      \ex{t{:}} & \ex{:client/\{ssn\} :address :location/\{ccity\}/\{cstreet\}} .
    \end{tabular}
  &
    {\sf\begin{tabular}{ll}
      :address a owl:ObjectProperty . \\
      :address rdfs:domain :Client . &
      :address rdfs:range :Location .
    \end{tabular}} \\
  \bottomrule
\end{tabular}
}
\end{center}
\textsl{References}: The pattern scenario foresees a separate table for the relationship. When this is the case, other approaches often prefer to apply a \emph{relationship reification} instead, where the relationship is reified into a class.
\begin{compactitem}
\item It is not present in \dm, where the scenario captured by \pat{SR} is instead handled through relationship reification (see \pat{SRR} pattern).
\item \pat{SR} slightly corresponds to the handling of binary relations in~\cite{SequedaArenasDM}, however that approach is substantially more limited than \pat{SR}:
  \begin{compactitem}
  \item It only applies to tables with exactly two attributes;
  \item The primary key must comprise both attributes;
  \item Incoming foreign keys are not allowed.
  \end{compactitem}
  On the other hand, the approach in~\cite{SequedaArenasDM} for binary relations applies also in those cases where the foreign keys do not refer the primary keys of the tables participating in the relationship. In our methodology, we decided to render this variant explicit through the dedicated Pattern~\pat{SRa}. Finally, their definition of foreign-key ignores the standard assumption that the referred attributes must form a key.
\item Case 2 in BootOX~\cite{JKZH*15} is comparable to \pat{SR}, however there are some ambiguities arising from the lack of a fixed conceptual representation. For instance, the treatment of BootOX is incomplete w.r.t.\ mandatory participation: constraints are handled correctly for role $R_E$, but they are ignored for role $R_F$. Other details of BootOX are omitted, and so it is unclear how they are handled: for instance, it is stated that mappings are generated according to \dm, however \dm does not handle the case of binary relationships (as discussed above), which is an apparent contradiction. BootOX does not impose any constraint on the attributes referred by the foreign keys, which could in principle not form a key.
\item In the algorithm of MIRROR~\cite{mirror}, ontology axioms over object properties are never included, since that algorithm does not consider \owl and only works at the level of R2RML. Modulo this major difference, our \pat{SR} pattern is \emph{partially} covered by different cases there, which differentiate from each other depending on the cardinality constraints in the logical schema:
  \begin{compactitem}
  \item Cases 5b and 8 resemble pattern \pat{SR} with a $(1,1)$ or a $(1,N)$ participation on one of the two roles, and $(0, N)$ on the other. A notable difference, though, is that in \pat{SR} the identification constraint entailed by the $(1,1)$ cardinality is correctly translated into a key constraint in the logical schema, whereas it is ignored by MIRROR.
  \item Case 7 resembles \pat{SR} with mandatory participation on both roles. In our case, however, we can exploit \owlql to partially encode the mandatory participation, using suitable class equivalence axioms.
  \item The case where both roles of a relationship participate with cardinality $(0,N)$ is pruned on purpose from their catalogue, with the argumentation that ``primary keys must be defined not null and unique''. Clearly, the fact that primary keys must be not null and unique is completely independent from the fact of having $(0,N)$ relationships in a conceptual model. Hence, we are not able to actually reconstruct what the authors' intention here was. 
  \end{compactitem}
\item Pattern \pat{SR} strictly includes \emph{Pattern 12: Many to Many Table} in~\cite{sequeda2012}. In particular, \pat{SR} covers all possibile cardinality constraints, whereas Pattern 12 should cover only the many-to-many case (according to its name). However, such pattern is not formally specified, and the choice on where to apply it is totally left open to the domain expert. One could in fact apply it even to tables representing binary relationships with arbitrary cardinality constraints, simply by ignoring such cardinalities. Pattern 12 does not discuss ontology axioms, and it does not rule out the case where the relationship has attributes (which cannot be correctly handled through \pat{SR}, nor through Pattern 12, because it requires reification as we will describe for Pattern \pat{SRR}).
\item Pattern \pat{SR} strictly includes pattern \emph{Relationship: Many to Many} in~\cite{sequeda-book}, since such pattern is just a renaming of \emph{Pattern 12: Many to Many Table} from~\cite{sequeda2012}. This pattern does not rule out the case where the relationship has attributes, which in the scope of that work is perfectly reasonable since the KGs they consider allow for specifying attributes on properties. This is common, for instance, in \emph{property graphs}~\cite{2021Hogan}. Recently, an extension of RDF called \emph{RDF-star}\footnote{\url{https://www.w3.org/2021/12/rdf-star.html}.} is being proposed that also incorporates this feature.
\end{compactitem}

\mypar{Schema Relationship with Identifier Alignment (SRa)}%
This pattern is similar to pattern \pat{SR}, but it comes with a \emph{modifier \pat{a}} indicating that the pattern can be applied after the identifiers involved in the relationship have been \emph{aligned}. The alignment is necessary because the foreign key in $T_R$ does not refer to the primary key $\sK_F$ of $T_F$, but to an alternative key $\sU_F$. Since the instances of the class $C_F$ corresponding to $T_F$ are constructed using the primary key $\sK_F$ of $T_F$ (cf.\ pattern \pat{SE}), also the pairs that populate $p_R$ should refer in their object position to that primary key, which can only be retrieved via a join between $T_R$ and $T_F$ on the non-primary key $\sU_F$.

Note that alignment variants can be defined in a straightforward way for other
patterns involving relationships. For conciseness, we prune these variants from
our catalog.
\\
\textsl{Example}: The primary key of the table with locations is not given by
the city and street, which are used in the table that relates clients to their
addresses, but is given by the latitude and longitude of locations.
\begin{center}
\noindent \resizebox{.9\textwidth}{!}{%
\begin{tabular}{c@{\hspace{2pt}}c}
  \toprule
  \textsc{Conceptual Model} & \textsc{DB schema}\\
  \multicolumn{2}{c}{(Datatypes Omitted)} \\
  \midrule
  \begin{minipage}{\erw}
    \centerline{\includegraphics[scale=1]{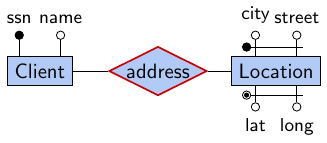}}
  \end{minipage}
  &
    \begin{tikzpicture}[dbpicture]
      \node (e)
      {\normalsize\ex{client}(\pk{\ex{s\tikz\coordinate(kE);sn}},\ex{name})};
      \node[right=of e] (f)
      {\normalsize\ex{location}(\ex{city\tikz\coordinate(kF);, street, \pk{lat, long}})} ;
      \node[below=of e.south west,anchor=north west] (r)
      {\normalsize\ex{address}(\pk{\ex{cli\tikz\coordinate(kRE);ent, ccity,\tikz\coordinate(kRF); cstreet}})};
      \draw[->] ($(kRE)+(0,\uoffset)$) -- ($(kRE)+(0,\uuoffset)$) -| ($(kE)-(0,\doffset)$);
      \draw[->] ($(kRF)+(0,\uoffset)$) |- ($(kRF)+(0,\uuoffset)$) -|
      ($(kF)-(0,\doffset)$);
      \node[above=of e, xshift=2cm] {\normalsize$\key[\ex{location}]{\ex{city,street}}$} ;
    \end{tikzpicture}
  \\[2em]
  \midrule
  \textsc{Mappings} & \textsc{Ontology} \\
  \midrule
  \begin{tabular}{l@{~}l}
    \ex{s{:}} & \ex{SELECT A.client, L.lat, L.long FROM address A, location L} \\
              & \ex{WHERE L.city=A.ccity AND L.street=A.cstreet} \\
    \ex{t{:}} & \ex{:client/\{ssn\} :address :location/\{lat\}/\{long\}} .
    \end{tabular}
  &
    {\sf\begin{tabular}{ll}
      :address a owl:ObjectProperty . \\
      :address rdfs:domain :Client . &
      :address rdfs:range :Location .
    \end{tabular}} \\
  \bottomrule
\end{tabular}
}
\end{center}
\textsl{References}: This pattern is original, as either it has never been considered by other approaches or it is mixed with the general strategy for handling binary relationships.
\begin{compactitem}
\item It is not present in \dm, where the scenario captured by \pat{SR} is instead handled through relationship reification.
\item \pat{SRa} covers the handling of binary relations in~\cite{SequedaArenasDM} for the case when the foreign keys does refer to non-primary keys of the participating entities. The issues of their approach highlighted in comparison to \pat{SR} apply for \pat{SRa} as well.
\item  \pat{SRa} covers the handling of binary relations in BootOX~\cite{JKZH*15} for the case when the foreign keys does refer to non-primary keys of the participating entities. The issues of their approach highlighted in comparison to \pat{SR} apply for \pat{SRa} as well.
\item \pat{SRa} does not correspond to any of the cases treated in MIRROR~\cite{mirror}.
\item \pat{SRa} does not correspond to any of the patterns presented in~\cite{sequeda-book}.
\end{compactitem}

\mypar{Schema Relationship with Merging (SRm)}  %
This pattern handles the case where the binary relationship is not rendered in the DB as a separate table, but rather merged into the table representing one of the participating entities, which can be done without introducing redundancy whenever such participation is of kind $(\_, 1)$. It
considers a table $T_E$ in which the foreign key
$\sK_{EF}$ referring a table $T_F$ is disjoint from its primary key $\sK_E$.  The
table $T_E$ is mapped to an object property, whose subject and object are
derived respectively from $\sK_E$ and $\sK_{EF}$.
\\
\textsl{Example}: The relationship between a client and its \emph{unique} billing
address is merged into the client table. In the ontology, a
\ex{billingAddress} object property relates the \ex{Client} class
to the \ex{Location} class, and is populated via a mapping from the client table.
\begin{center}
\noindent \resizebox{.9\textwidth}{!}{%
\begin{tabular}{c@{\hspace{2pt}}c}
  \toprule
  \textsc{Conceptual Model} & \textsc{DB schema}\\
    \multicolumn{2}{c}{(Datatypes Omitted)} \\
  \midrule
  \begin{minipage}{\erw}
    \centerline{\includegraphics[scale=1]{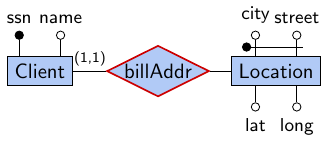}}
  \end{minipage}
  &
    \begin{tikzpicture}[dbpicture]
      \node (e) at (0,.3)
      {\normalsize\ex{client}(\pk{\ex{s\tikz\coordinate(kE);sn}},\ex{name}, ccity\tikz\coordinate(kRE);, cstreet)};
      \node[below=of e.south west,anchor=north west] (f)
      {\normalsize\ex{location}(\ex{\pk{city\tikz\coordinate(kF);, street}, lat, long})} ;
      \draw[->] ($(kRE)+(0,-\doffset)$) -- ($(kRE)+(0,-\uoffset)$) -| ($(kF)-(0,-\uoffset)$);
    \end{tikzpicture}
  \\[2em]
  \midrule
  \textsc{Mappings} & \textsc{Ontology} \\
  \midrule
  \begin{tabular}{l@{~}l}
    \ex{s{:}} & \ex{SELECT ssn, ccity, cstreet FROM client} \\
    \ex{t{:}} & \ex{:client/\{ssn\} :billingAddress :location/\{ccity\}/\{cstreet\}} .
    \end{tabular}
  &
    {\sf\begin{tabular}{ll}
      :billingAddress a owl:ObjectProperty . \\
      :billingAddress rdfs:domain :Client . &
      :billingAddress rdfs:range :Location . \\
      :Client rdfs:subClassOf _:r1 . &
      _:r1 a owl:Restriction . \\
      _:r1 owl:onProperty :billingAddress . &
      _:r1 owl:someValuesFrom owl:Thing . \\
        \end{tabular}} \\
  \bottomrule
\end{tabular}
}
\end{center}
\textsl{References}: This pattern, as its alignment variant \pat{SRma}, slightly overlaps with the common strategy of transforming each foreign key into an object property.
\begin{compactitem}
\item It is partially covered by \dm, where each foreign key is translated to an object property. However, there are substantial differences, going beyond the fact that \owlql constraints are not considered in \dm. For instance, in case of $(1,1)$ cardinality, where the foreign-key holds in both directions, \dm prescribes the creation of two different object properties, whereas \pat{SRm} would create only one property (the one focused on the relationship), encoding the additional foreign-key as an additional \owlql inclusion.
\item Since~\cite{SequedaArenasDM} adopts an approach close to \dm, similar considerations apply.
\item It roughly corresponds to \emph{Case 4} in BootOX~\cite{JKZH*15}. However, since the catalog of BootOX does not fix the underlying conceptual representation, ambiguities can arise: for instance, \pat{SRm} adds a mandatory participation constraint only if such constraint derives from a mandatory participation in the ER-diagram, whereas constraints in BootOX are explicitly handled by the user (without an indication on how this should be done). Similarly to what we observed when discussing Pattern~\pat{SR}, BootOX Case 4  seems to be incomplete, for example, cardinality constraints on the role for the entity $F$ are ignored.
\item In the algorithm of MIRROR~\cite{mirror}, a substantial portion of \pat{SRm} corresponds to the combination of cases 2b, 4, and 5a. The only difference is that pattern \pat{SRm} also covers the case when $F$ participates with cardinality $(1,1)$ to $R$, and that MIRROR does not produce ontology axioms by design.
\item In~\cite{sequeda-book}, the only pattern resembling \pat{SRm} is ``\emph{Direct Relationship}'', which is the same approach used in \dm. Hence, all considerations discussed for \dm apply to \emph{Direct Relationship} as well.
\end{compactitem}

\mypar{Schema Weak-Entity (\pat{SEw})}  %
This pattern considers a \emph{weak entity} $E$ identified through a relationship $R$. The
table $T_E$, corresponding to $E$, is mapped to a class $C_E$, whose instances are built through the primary identifier of $T_E$, and whose data properties correspond to the attributes of $E$.
The relationship $R$ is captured through an object property $p_{EF}$, as in \pat{SRm}.
\\
\textsl{Example}: A room in a university is identified by a code and the building it belongs to. Since a room must belong to exactly one building, the relationship \ex{belongsTo} has been merged into entity \ex{Room}, and \ex{Room} is a weak entity identified through \ex{belongsTo}. In the ontology, a class \ex{Room} is created, populated via a mapping that builds \ex{Room} individuals by combining the code of the room together with the identifier of the building. \ex{Room} individuals are put in correspondence to their respective \ex{Building} individuals through an appropriate \ex{belongsTo} object property.
\begin{center}
\noindent \resizebox{.9\textwidth}{!}{%
\begin{tabular}{c@{\hspace{2pt}}c}
  \toprule
  \textsc{Conceptual Model} & \textsc{DB schema}\\
  \multicolumn{2}{c}{(Datatypes Omitted)} \\
  \midrule
  \begin{minipage}{\erw}
    \centerline{\includegraphics[scale=1]{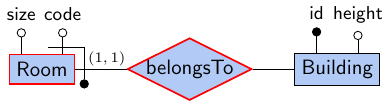}}
  \end{minipage}
                            &
    \begin{tikzpicture}[dbpicture]
        \node[yshift=.3cm] (f)
          {\normalsize\ex{Building(\pk{i\tikz\coordinate(kF);d}, height)}};
        \node[below=of f.south west,anchor=north west] (e)
          {\normalsize\ex{Room(\pk{code,buil\tikz\coordinate(kEF);ding},height)}};
        \draw[->] ($(kEF)+(0,\uoffset)$) |- ($(kEF)+(0,\uuoffset)$) -|
          ($(kF)-(0,\doffset)$);
        \end{tikzpicture}
  \\[2em]
  \midrule
  \textsc{Mappings} & \textsc{Ontology} \\
  \midrule
  \begin{tabular}{l@{~}l}
    \ex{s{:}} & \ex{SELECT * FROM building} \\
    \ex{t{:}} & \ex{:room/\{code\}/\{building\} a :Room} . \\
              & \ex{:room/\{code\}/\{building\} :room\#code \{code\}} .\\
              & \ex{:room/\{code\}/\{building\} :room\#size \{size\}} .\\
              & \ex{:room/\{code\}/\{building\} :belongsTo :Building/\{building\}} .
    \end{tabular}
  &
    {\sf\begin{tabular}{ll}
     :Room a owl:Class . & :room\#code a owl:DatatypeProperty . \\
      :room\#size a owl:DatatypeProperty . & :belongsTo a owl:ObjectProperty . \\
      :room\#code rdfs:domain :Client . &
      :room\#size rdfs:domain :Client . \\
      :belongsTo rdfs:domain :Room . & :belongsTo rdfs:range :Client . \\
      :Room rdfs:subClassOf _:r1 . &
      _:r1 a owl:Restriction . \\
      _:r1 owl:onProperty :room\#code . &
      _:r1 owl:someValuesFrom rdfs:Literal . \\
      :Client rdfs:subClassOf _:r2 . &
      _:r2 a owl:Restriction . \\
      _:r2 owl:onProperty :room\#size . &
      _:r2 owl:someValuesFrom rdfs:Literal . \\
      :Client rdfs:subClassOf _:r3 . &
      _:r3 a owl:Restriction . \\
      _:r3 owl:onProperty :belongsTo . &
      _:r3 owl:someValuesFrom owl:Thing .
    \end{tabular}} \\
  \bottomrule
\end{tabular}
}
\end{center}
\textsl{References}: This pattern, and its alignment variant \pat{SEwa}, slightly overlaps with the common strategy of transforming each foreign key into an object property.
\begin{compactitem}
\item It is partially covered by \dm, where each foreign key is translated to an object property, and object identifiers are always built from primary keys. However, the differences already mentioned for \pat{SE} and \pat{SRm} apply to this pattern as well.
\item Since~\cite{SequedaArenasDM} adopts an approach close to \dm, similar considerations apply.
\item From a DB schema as in \pat{SEw}, the combination of Rules 1, 2, 3, 4 and 7 from BootOX~\cite{JKZH*15} would generate mappings and ontology conforming to \pat{SEw}.
\item \pat{SEw} is partially captured by Case 6b of MIRROR~\cite{mirror}. MIRROR only captures the case of $(0,N)$ cardinality for role $R_F$, whereas \pat{SEw} captures all possible cases. Another difference is that, in MIRROR, no alignment variant \pat{SEwa} is discussed. Finally, MIRROR does not produce ontology axioms by design.
\item To obtain mapping assertions of \pat{SEw} from~\cite{sequeda-book}, one can apply the ``Direct'' Patterns described therein, which correspond to \dm. Such work does not consider ontology axioms.
\end{compactitem}

\mypar{Schema Reified Relationship (SRR)} %
This pattern deals with n-ary relationships and/or relationships with attributes. For both
cases, it is necessary to \emph{reify} the relationship into a class first. This is because RDF
can only encode unary predicates (through class assertions) and binary predicates (through
object property assertions), and it does not allow for attributes over properties (recall that this is instead possible in property graphs~\cite{2021Hogan}).
The pattern considers a table $T_R$ whose primary key is partitioned in at
least three parts $\sK_{RE}$, $\sK_{RF}$, and $\sK_{RG}$, that are foreign keys
to three additional tables; or when the primary key is partitioned in at least
two such parts, but there are additional attributes in $T_R$.
Such a table naturally corresponds to an $n$-ary relationship $R$
with $n > 2$ (or with attributes), and to represent it at the
ontology level we require a class $C_R$, which reifies $R$,
whose instances are built
from the primary key of $T_R$. The mapping
accounts for the fact that the components of the $n$-ary relationship
have to be represented by suitable object properties, one for each such
component, and that the tuples that instantiate these object properties can all
be derived from $T_R$ alone.
\\
\textsl{Example}: A table containing information about signed contracts, which
involve a player, a team, and the contracts themselves. This
information is represented by a relationship that is inherently ternary.
The ontology should contain a class corresponding to the reification of such relationship, e.g., a class \ex{Signs}.
\begin{center}
\resizebox{.9\textwidth}{!}{%
\begin{tabular}{c@{\hspace{2pt}}c}
  \toprule
  \textsc{Conceptual Model} & \textsc{DB schema}\\
  \multicolumn{2}{c}{(Datatypes Omitted)} \\
  \midrule
  \begin{minipage}{\erw}
    \centerline{\includegraphics[scale=1]{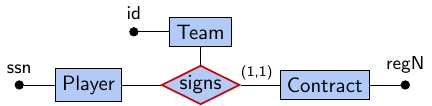}}
  \end{minipage}
                            &
      \begin{tikzpicture}[dbpicture]
        \node[yshift=.6cm] (g)
          {\normalsize team(\pk{i\tikz\coordinate(kG);d})};
        \node[below=of g] (r)
          {\normalsize signs(\pk{pla\tikz\coordinate(kRE);yer,
            cont\tikz\coordinate(kRF);ract,
            te\tikz\coordinate(kRG);am})};
        \node[below=of r.south west,anchor=north west] (e)
          {\normalsize player(\pk{s\tikz\coordinate(kE);sn})};
        \node[right=of e] (f)
          {\normalsize contract(\pk{re\tikz\coordinate(kF);gN})};
        \draw[->] ($(kRE)-(0,\doffset)$) -| ($(kRE)-(0,\ddoffset)$) -| ($(kE)+(0,\uoffset)$);
        \draw[->] ($(kRF)-(0,\doffset)$) |- ($(kRF)-(0,\ddoffset)$) -|
          ($(kF)+(0,\uoffset)$);
        \draw[->] ($(kRG)+(0,\uoffset)$) |- ($(kRG)+(0,\uuoffset)$) -|
          ($(kG)-(0,\doffset)$);
        \end{tikzpicture}
  \\[2em]
  \midrule
  \textsc{Mappings} & \textsc{Ontology} \\
  \midrule
  \begin{tabular}{l@{~}l}
    \ex{s{:}} & \ex{SELECT * FROM signs} \\
    \ex{t{:}} & \ex{:signs/\{player\}/\{contract\}/\{team\} a :Signs} . \\
              & \ex{:signs/\{player\}/\{contract\}/\{team\}} \\
              & \qquad\qquad \ex{:hasContract :Contract/\{contract\}} . \\
              & \ex{:signs/\{player\}/\{contract\}/\{team\}} \\
              & \qquad\qquad\ex{:hasPlayer :Player/\{player\}} .\\
              & \ex{:signs/\{player\}/\{contract\}/\{team\}} \\
              & \qquad\qquad\ex{:hasTeam :Team/\{team\}} .\\
    \end{tabular}
  &
    {\sf\begin{tabular}{ll}
      :Signs a owl:Class . & :hasContract a owl:ObjectProperty . \\
      :hasPlayer a owl:ObjectProperty . & :hasTeam a owl:ObjectProperty . \\
      :hasContract rdfs:domain :Signs . &
      :hasContract rdfs:range :Contract . \\
      :hasPlayer rdfs:domain :Signs . &
      :hasPlayer rdfs:range :Player .\\
      :hasTeam rdfs:domain :Signs . &
      :hasTeam rdfs:range :Team . \\
      :Signs rdfs:subClassOf _:r1 . &
      _:r1 a owl:Restriction . \\
      _:r1 owl:onProperty :hasContract . &
      _:r1 owl:someValuesFrom owl:Thing . \\
      :Signs rdfs:subClassOf _:r2 . &
      _:r2 a owl:Restriction . \\
      _:r2 owl:onProperty :hasPlayer . &
      _:r2 owl:someValuesFrom owl:Thing . \\
      :Signs rdfs:subClassOf _:r3 . &
      _:r3 a owl:Restriction . \\
      _:r3 owl:onProperty :hasTeam . &
      _:r3 owl:someValuesFrom owl:Thing . \\
      :Contract rdfs:subClassOf _:r4 . &
      _:r4 a owl:Restriction . \\
      _:r4 owl:onProperty _:r5 . &
      _:r5 owl:inverseOf :hasContract . \\
      _:r4 owl:someValuesFrom owl:Thing . \\
    \end{tabular}} \\
  \bottomrule
\end{tabular}
}
\end{center}
\textsl{References}: This pattern, which corresponds to \emph{reification} in ontological and conceptual modeling~\cite{CaLN98,CaLN99,BeCD05}, is commonly used to handle the case of ``many-to-many'' tables. A main difference with respect to other approaches is that we do not create data properties for the attributes identifying the reified relationship (i.e., entity $R$), because such attributes are already being represented as data properties of the classes encoding the entities participating to the relationship.
\begin{compactitem}
\item In \dm, every table encoding a relationship is handled according to a strategy similar to \pat{SRR}, devoid of ontology axioms.
\item Also~\cite{SequedaArenasDM} adopts an approach very similar to \dm, so the same considerations discussed for \dm apply as well, apart from the fact that domain and range axioms (but not mandatory participations) are added to the ontology.
\item Case (2) of BootOX~\cite{JKZH*15} is used to handle both the case of binary relationships and general relationships needing reification. For these reasons, the same considerations already discussed for \pat{SR} apply to \pat{SRR} as well.
\item None of the cases in MIRROR~\cite{mirror} handles the situation of a table encoding an n-ary relationship or a relationship with attributes.
\item The notion of reification is also present in the ``Direct'' Patterns of~\cite{sequeda-book}, given that such patterns encode \dm. A notable difference, though, is that the focus there is on property graphs, which as mentioned allow to avoid reification for the case of binary relationship with attributes.
\end{compactitem}

\mypar{Schema Hierarchy (SH)}
This pattern captures the most common case of ISA (i.e., the parent-child E-R relation between entities) between entities. It considers a table $T_F$ whose primary key is a foreign key referring the primary key of a table $T_E$. Then, $T_F$ is mapped to a class $C_F$ in the ontology that is a sub-class of the class $C_E$ to which $T_E$ is mapped. Hence, $C_F$ ``inherits'' the template $\iri_E$ of $C_E$, so that the instances of the two classes are ``compatible''. Note that we here discuss the case where both the child and the parent tables are maintained in the database schema. Other strategies for handling ISAs between entities might be considered as well~\cite{HaMo10}.
\\
\textsl{Example}: An entity Student in an ISA relation with an entity Person.
\begin{center}
\noindent \resizebox{.9\textwidth}{!}{%
\begin{tabular}{c@{\hspace{2pt}}c}
  \toprule
  \textsc{Conceptual Model} & \textsc{DB schema}\\
  \multicolumn{2}{c}{(Datatypes Omitted)} \\
  \midrule
  \begin{minipage}{\erw}
    \centerline{\includegraphics[scale=1]{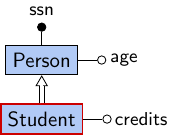}}
  \end{minipage}
  &
    \begin{tabular}{l}
      \begin{tikzpicture}[dbpicture]
        \node (e)
        {\normalsize person(\pk{ss\tikz\coordinate(kE);n},age)};
        \node[below=of e.south west,anchor=north west] (f)
        {\normalsize student(\pk{ss\tikz\coordinate(kF);sn},credits)};
        \draw[->] ($(kF)+(0,\uoffset)$) -| ($(kF)+(0,\uuoffset)$) -| ($(kE)-(0,\doffset)$);
      \end{tikzpicture}
    \end{tabular}
  \\[2em]
  \midrule
  \textsc{Mappings} & \textsc{Ontology} \\
  \midrule
  \begin{tabular}{l@{~}l}
      \ex{s{:}} & \ex{SELECT * FROM student} \\
      \ex{t{:}} & \ex{:person/\{sssn\} a :Student} . \\
                & \ex{:person/\{sssn\} :student\#credits \{credits\}} .\\
    \end{tabular}
  &
    \begin{tabular}{ll}
      :Student a owl:Class . & :Student rdfs:subClassOf :Person . \\
      :student\#credits a owl:DatatypeProperty . &
      :student\#credits rdfs:domain :Student . \\
      :Student rdfs:subClassOf _:r1 . &
      _:r1 a owl:Restriction . \\
      _:r1 owl:onProperty :student\#credits . &
      _:r1 owl:someValuesFrom rdfs:Literal . \\
    \end{tabular} \\
  \bottomrule
\end{tabular}
}
\end{center}
\textsl{References}: This is an advanced pattern, and is present only in few approaches.
\begin{compactitem}
\item It is not present in \dm, where each foreign-key is translated into an object property instead.
\item It is also not present in the extension of \dm discussed in~\cite{SequedaArenasDM}.
\item Class subsumption is considered in Case (5) of BootOX~\cite{JKZH*15}. However, it is unclear how the mapping assertions are actually built. In the text, it is stated that mappings are generated according to \dm, however such a strategy would not produce the desired results, since the templates used for generating objects of the subclass differ from those used for generating objects of the superclass.
\item Case 2b of MIRROR~\cite{mirror} should handle ISAs between two entities, however it differs from \pat{SH} and BootOX~\cite{JKZH*15} since, in the DB schema, the foreign key is not required to be referring to a key. This seems to be a glitch.
\item Class hierarchies are not discussed in any of the mapping patterns from~\cite{sequeda-book}.
\end{compactitem}

\mypar{Schema Hierarchy with Identifier Alignment (SHa)}
Such pattern is like \pat{SH}, apart from the foreign-key constraint that can come in three different variants. In the depicted one, the foreign key in $T_F$ is over a non-primary key $\sK_{FE}$. The objects for $C_F$ have to be built out of $\sK_{FE}$, rather than out of its primary key. For this purpose, the pattern creates a view $V_F$ in which $\sK_{FE}$ is the primary key, and the foreign key relations are preserved. Such view might enable further applications of patterns (see Example~\ref{e:example-schema-driven}).
\\
\textsl{Example}: An ISA relation between entities Student and Person. Students are identified by their matriculation number, whereas persons are identified by their SSN.
\begin{center}
\noindent \resizebox{.9\textwidth}{!}{%
\begin{tabular}{c@{\hspace{2pt}}c}
  \toprule
  \textsc{Conceptual Model} & \textsc{DB schema}\\
  \multicolumn{2}{c}{(Datatypes Omitted)} \\
  \midrule
  \begin{minipage}{\erw}
    \centerline{\includegraphics[scale=1]{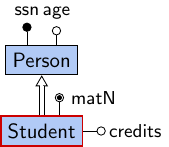}}
  \end{minipage}
  &
    \begin{tabular}{l}
      \begin{tikzpicture}[dbpicture]
        \node (e)
        {\normalsize person(\pk{ss\tikz\coordinate(kE);n},age)};
        \node[above=of e, xshift=0cm] {\normalsize$\key[\ex{student}]{\ex{sssn}}$} ;
        \node[below=of e.south west,anchor=north west] (f)
        {\normalsize student(ss\tikz\coordinate(kF);sn,\pk{matN},credits)};
        \draw[->] ($(kF)+(0,\uoffset)$) -| ($(kF)+(0,\uuoffset)$) -| ($(kE)-(0,\doffset)$);
      \end{tikzpicture}
    \end{tabular}
  \\[2em]
  \midrule
  \textsc{Mappings} & \textsc{Ontology} \\
  \midrule
  \begin{tabular}{l@{~}l}
      \ex{s{:}} & \ex{SELECT * FROM student} \\
      \ex{t{:}} & \ex{:person/\{sssn\} a :Student} . \\
                & \ex{:person/\{sssn\} :student\#matN \{matN\}} .\\
                & \ex{:person/\{sssn\} :student\#credits \{credits\}} .\\
    \end{tabular}
  &
    \begin{tabular}{ll}
      :Student a owl:Class . & :Student rdfs:subClassOf :Person . \\
      :student/\#matN a owl:DatatypeProperty . &
      :student/\#credits  a owl:DatatypeProperty . \\
      :student\#credits rdfs:domain :Student . &
      :student\#matN rdfs:domain :Student . \\
      :Student rdfs:subClassOf _:r1 . &
      _:r1 a owl:Restriction . \\
      _:r1 owl:onProperty :student\#credits . &
      _:r1 owl:someValuesFrom rdfs:Literal . \\
      :Student rdfs:subClassOf _:r2 . &
      _:r2 a owl:Restriction . \\
      _:r2 owl:onProperty :student\#matN . &
      _:r2 owl:someValuesFrom rdfs:Literal . \\
    \end{tabular} \\
  \bottomrule
\end{tabular}
}
\end{center}
\textsl{References}: This is an advanced pattern, and is present only in few approaches.
\begin{compactitem}
\item It is not present in \dm, where each foreign-key is translated into an object property instead.
\item It is not present in~\cite{SequedaArenasDM}.
\item Case (5) of BootOX~\cite{JKZH*15} also deals also with the situation in which the foreign-key points to a non-primary identifier of the parent entity. However, it does not deal with the two other variants admitted by \pat{SHa}. Regarding the mapping assertions, the same considerations discussed for \pat{SH} apply here as well.
\item Regarding MIRROR~\cite{mirror}, the same considerations we had for \pat{SH} apply.
\item Class hierarchies are not discussed in any of the mapping patterns from~\cite{sequeda-book}.
\end{compactitem}

We now provide an example showing one of the possible usages of schema-driven mapping patterns, specifically, to derive an ontology and mappings starting from a conceptualization and a DB schema.

\begin{figure}
\centering
\begin{tabular}{c}
      \begin{tabular}{cc}
        \begin{minipage}{.3\linewidth}
          \includegraphics[width=\textwidth]{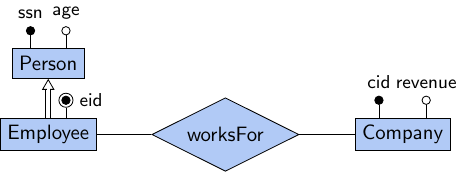}
        \end{minipage}
        &
          \begin{minipage}{.4\linewidth}
            \resizebox{\textwidth}{!}{
              \begin{tikzpicture}[dbpicture]
                \sf
                \node (e) at (-1,0)
                {Person(\pk{ssn\tikz\coordinate(ssn);},age)};
                \node (f) at (3,0)
                {Company(\pk{cid\tikz\coordinate(cid);},revenue)};
                \node (r) at (1.5,-.75)
                {worksFor(\pk{weid\tikz\coordinate(weid);, wcid\tikz\coordinate(wcid);})};
                \node (Er) at (0,-1.5) {
                  Employee(ssn\tikz\coordinate(essn);, \pk{eid}\tikz\coordinate(eid);, age)
                };
                \draw[->] ($(weid)+(-2mm,-1mm)$) |- ($(weid)-(2mm,\ddoffset)$) -| ($(eid)-(1mm,-\uoffset)$);
                \draw[->] ($(wcid)+(-2mm,\uoffset)$) |- ($(wcid)+(-2mm,\ddoffset)$) -| ($(cid)-(1.5mm,1mm)$);
                \draw[->] ($(essn) + (-2mm,\uoffset)$) |- ($(ssn) + (-1.9mm,-\uuoffset)$) |- ($(ssn) + (-1.9mm,-\doffset)$);
                \node (key) at (3,-1.5) {$\key[\sf{Employee}]{\sf ssn}$} ;
              \end{tikzpicture}
            }
          \end{minipage}
      \end{tabular}\\
      \midrule
      \begin{tabular}{cc}
        \begin{minipage}{.3\linewidth}
          \includegraphics[width=\textwidth]{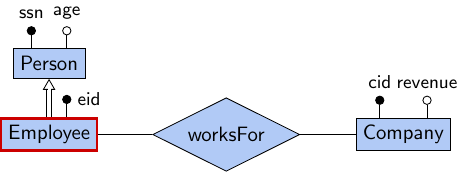}
        \end{minipage}
        &
          \begin{minipage}{.4\linewidth}
            \resizebox{\textwidth}{!}{
              \begin{tikzpicture}[dbpicture]
                \sf
                \node (e) at (-1,0)
                {Person(\pk{ssn\tikz\coordinate(ssn);},age)};
                \node (f) at (3,0)
                {Company(\pk{cid\tikz\coordinate(cid);},revenue)};
                \node (r) at (1.5,-.75)
                {worksFor(\pk{weid\tikz\coordinate(weid);, wcid\tikz\coordinate(wcid);})};
                \node (Er) at (0,-1.5) {
                  $\sf{V_{Employee}}$(\pk{ssn}\tikz\coordinate(essn);, eid\tikz\coordinate(eid);,age)
                };
                \draw[->] ($(weid)+(-2mm,-1mm)$) |- ($(weid)-(2mm,\ddoffset)$) -| ($(eid)-(0.7mm,-\uoffset)$);
                \draw[->] ($(wcid)+(-2mm,\uoffset)$) |- ($(wcid)+(-2mm,\ddoffset)$) -| ($(cid)-(1.5mm,1mm)$);
                \draw[->] ($(essn) + (-2mm,\uoffset)$) |- ($(ssn) + (-1.9mm,-\uuoffset)$) |- ($(ssn) + (-1.9mm,-\doffset)$);
                \node (key) at (3,-1.5) {$\key[\sf{V_{Employee}}]{\sf eid}$} ;
              \end{tikzpicture}
            }
          \end{minipage}
      \end{tabular}\\
          \midrule
      \begin{tabular}{cc}
        \begin{minipage}{.3\linewidth}
          \includegraphics[width=\textwidth]{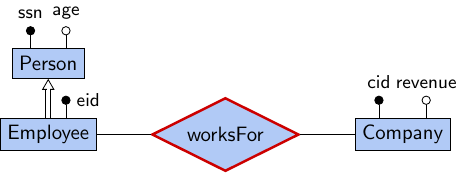}
        \end{minipage}
        &
          \begin{minipage}{.4\linewidth}
            \resizebox{\textwidth}{!}{
              \begin{tikzpicture}[dbpicture]
                \sf
                \node (e) at (-1,0)
                {Person(\pk{ssn\tikz\coordinate(ssn);},age)};
                \node (f) at (3,0)
                {Company(\pk{cid\tikz\coordinate(cid);},revenue)};
                \node (r) at (1.5,-.75)
                {$\sf V_{worksFor}$(\pk{ssn\tikz\coordinate(wssn);, wcid\tikz\coordinate(wcid);})};
                \node (Er) at (0,-1.5) {
                  $\sf{V_{Employee}}$(\pk{ssn}\tikz\coordinate(essn);, eid\tikz\coordinate(eid);, age)
                };
                \draw[->] ($(wssn)+(-2mm,-1mm)$) |- ($(wssn)-(2mm,\ddoffset)$) -| ($(essn)-(0.7mm,-\uoffset)$);
                \draw[->] ($(wcid)+(-2mm,\uoffset)$) |- ($(wcid)+(-2mm,\ddoffset)$) -| ($(cid)-(1.5mm,1mm)$);
                \draw[->] ($(essn) + (-3mm,\uoffset)$) |- ($(ssn) + (-1.9mm,-\uuoffset)$) |- ($(ssn) + (-1.9mm,-\doffset)$);
                \node (key) at (3,-1.5) {$\key[\sf{V_{Employee}}]{\sf eid}$} ;
              \end{tikzpicture}
            }
          \end{minipage}
      \end{tabular}
\end{tabular}
  \caption{\label{f:example-schema-driven} The application of patterns can induce a refactoring of the DB schema and conceptual model.}
\end{figure}

\begin{example}\label{e:example-schema-driven}
  Consider situation depicted in the top row of Figure~\ref{f:example-schema-driven}. For conciseness, we omit datatypes. We use schema-driven patterns to derive an ontology and mappings.

  Under such a configuration of conceptual model/DB schema, we can only apply pattern \pat{SE} on \ex{Company} or \ex{Person}. As URI template functions, we here adopt the W3C Direct Mapping convention, assuming the \emph{base URI} \texttt{http://www.example.com/}\footnote{Hence, the prefix ``:'' is associated to the URI \texttt{http://www.example.com/}}.

  We start with the entity \ex{Company}. The application of \pat{SE} yields the following mapping assertion and set of ontology axioms:
  \begin{center}
    \resizebox{.9\textwidth}{!}{%
\begin{tabular}{c@{\hspace{2pt}}c}
  \toprule
  \textsc{Mappings} & \textsc{Ontology} \\
  \multicolumn{2}{c}{(Datatypes and Mandatory Participations Omitted)} \\
  \midrule
  \sf
  \begin{tabular}{l@{~}l}
      \ex{s{:}} & \ex{SELECT * FROM Company} \\
      \ex{t{:}} & \ex{:Company/cid=\{cid\} a :Company} ; \\
                & \ex{\qquad :Company\#cid \{cid\}} ;\\
                & \ex{\qquad :Company\#revenue \{revenue\}} .\\
    \end{tabular}
  &\sf
    \begin{tabular}{ll}
      :Company a owl:Class . & :Company\#cid a owl:DatatypeProperty .\\
      :Company\#revenue a owl:DatatypeProperty . & :Company\#cid rdfs:domain :Company . \\
      :Company\#revenue rdfs:domain :Company . \\
    \end{tabular} \\
  \bottomrule
\end{tabular}
}
\end{center}
We proceed similarly for entity \ex{Person}:
\begin{center}
  \resizebox{.9\textwidth}{!}{%
\begin{tabular}{c@{\hspace{2pt}}c}
  \toprule
  \textsc{Mappings} & \textsc{Ontology} \\
  \multicolumn{2}{c}{(Datatypes and Mandatory Participations Omitted)} \\
  \midrule
  \sf
  \begin{tabular}{l@{~}l}
      \ex{s{:}} & SELECT * FROM Person \\
      \ex{t{:}} & :Person/ssn=\{ssn\} a :Person ; \\
                & \qquad :Person\#ssn \{ssn\}; \\
                & \qquad :Person\#age \{age\} .\\
    \end{tabular}
  &\sf
    \begin{tabular}{ll}
      :Person a owl:Class . \\
      :Person\#ssn a owl:DatatypeProperty . & :Person\#age a owl:DatatypeProperty . \\
      :Person\#ssn rdfs:domain :Person . & :Person\#age rdfs:domain :Person .
    \end{tabular} \\
  \bottomrule
\end{tabular}
}
\end{center}
Since the URI template for the superclass \ex{Person} has been established, Pattern \pat{SHa} becomes now applicable over entity \ex{Employee}:
\begin{center}
\resizebox{.9\textwidth}{!}{%
\begin{tabular}{c@{\hspace{2pt}}c}
  \toprule
  \textsc{Mappings} & \textsc{Ontology} \\
  \multicolumn{2}{c}{(Datatypes and Mandatory Participations Omitted)} \\
  \midrule
  \sf
  \begin{tabular}{l@{~}l}
      \ex{s{:}} & SELECT * FROM Employee \\
    \ex{t{:}} & :Person/ssn=\{ssn\} a :Employee ; \\
                & \qquad :Employee\#eid \{eid\} . \\
    \end{tabular}
  &\sf
    \begin{tabular}{ll}
      :Employee a owl:Class . \\
      :Employee\#eid a owl:DatatypeProperty . &
      :Employee\#eid rdfs:domain :Employee .
    \end{tabular} \\
  \bottomrule
\end{tabular}
}
\end{center}
As by-product of the application of such pattern, we also obtain the updated conceptual model and DB schema depicted in the middle row of Figure~\ref{f:example-schema-driven}. Such by-product enables the application of the \pat{SRa} over relationship \ex{worksFor}, leading to the following mapping assertion and ontology axioms:
\begin{center}
\resizebox{.9\textwidth}{!}{%
\begin{tabular}{c@{\hspace{2pt}}c}
  \toprule
  \textsc{Mappings} & \textsc{Ontology} \\
  \multicolumn{2}{c}{(Datatypes and Mandatory Participations Omitted)} \\
  \midrule
  \sf
  \begin{tabular}{l@{~}l}
    \ex{s{:}} & SELECT ssn, wcid FROM Employee \\
              & JOIN worksFor ON weid=eid \\
    \ex{t{:}} & :Person/ssn=\{ssn\} :worksFor/\{ssn\}/\{wcid\} ; \\
              & \qquad :Company/cid=\{wcid\} .
    \end{tabular}
  &\sf
    \begin{tabular}{ll}
      :worksFor a owl:ObjectProperty . \\
      :worksFor rdfs:domain :Employee . & :worksFor rdfs:range :Company .
    \end{tabular} \\
  \bottomrule
\end{tabular}
}
\end{center}
No further pattern is applicable, and the obtained ontology is indeed a DB ontology because it represents all the entities and relationships in the conceptual model.
\end{example}


%% file: tables/schema-driven-1.tex
\begin{table}[t!]
  \centering
  {\linespread{0.96}\selectfont
  \resizebox{\textwidth}{!}{%
  \begin{tabularx}{1.5\textwidth}{l@{\hspace{-2pt}}c@{\hspace{1pt}}ll@{}}
    \toprule
    \textsc{Conceptual Model} & \textsc{DB schema} & \textsc{Mappings} & \textsc{Ontology}\\
    \midrule
    \multicolumn{4}{c}{\textbf{Schema Entity (SE)}}\\
    \begin{minipage}{\erw}
      \centerline{\includegraphics[scale=\erscale]{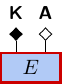}}
    \end{minipage}
    &
    $T_E(\pk{\sK},\sA)$
    &
    $\begin{array}{l@{~}l}
      s{:} & T_E \\
      t{:} & C_E(\iri_E(\sK)), \\
           & \{d_A(\iri_E(\sK),A)\}_{A \in \sK\cup\sA}
    \end{array}$
    &
    $\begin{array}{l@{}}
       \left\{\begin{array}{l}
                \delta(d_A) \sqsubseteq C_E, \\
                \rho(d_A) \sqsubseteq \mu(\tau(A)), \\
                C_E \sqsubseteq \delta(d_{A})
              \end{array}
       \right\}_{A \in \sK\cup\sA}
    \end{array}$
    \\[5ex]
    \multicolumn{4}{l}{\small
         \begin{minipage}{1.48\textwidth}
           \setdefaultleftmargin{1em}{0em}{0em}{0em}{0em}{0em}
           In case of optional attributes, for each optional attribute $A'$ add a $\nullable(A')$ constraint to the DB schema and drop the corresponding axiom $C_E \sqsubseteq \delta(d_{A'})$ from the ontology.
         \end{minipage}
       } \\
    \midrule
    \multicolumn{4}{c}{\textbf{Schema Relationship (SR)}}\\
    \begin{minipage}{\erw}
      \includegraphics[scale=\erscale]{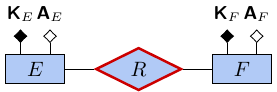}
    \end{minipage}
    &
    \begin{tabular}{l}
      \begin{tikzpicture}[dbpicture]
        \node (e)
          {$T_E(\pk{\sK\tikz\coordinate(kE);_E},\sA_E)$};
        \node[right=of e] (f)
          {$T_F(\pk{\sK\tikz\coordinate(kF);_F},\sA_F)$};
        \node[below=of e.south west,anchor=north west] (r)
          {$T_R(\pk{\sK\tikz\coordinate(kRE);_{RE},
              \sK\tikz\coordinate(kRF);_{RF}})$};
        \draw[->] ($(kRE)+(0,\uoffset)$) -- ($(kE)-(0,\doffset)$);
        \draw[->] ($(kRF)+(0,\uoffset)$) |- ($(kRF)+(0,\uuoffset)$) -|
          ($(kF)-(0,\doffset)$);
      \end{tikzpicture}
    \end{tabular}
    &
      $\begin{array}{l@{~}l}
        s{:} & T_R \\
        t{:} & p_R(\iri_{C_E}(\sK_{RE}),\iri_{C_F}(\sK_{RF}))
      \end{array}$
    &
      $\begin{array}{l@{}}
        \exists p_R \sqsubseteq C_E \\
        \exists p_R^- \sqsubseteq C_F
       \end{array}$\\[5ex]
       \multicolumn{4}{l}{\small
         \begin{minipage}{1.48\textwidth}
           \setdefaultleftmargin{1em}{0em}{0em}{0em}{0em}{0em}
           \begin{compactitem}
             \item In case of $(\_,1)$ cardinality on role $R_E$ (resp., $R_F$), the primary key for $T_R$ is restricted to the attributes $\sK_{RE}$ (resp., $\sK_{RF}$). In case both roles have $(\_,1)$ cardinality, either choice for the primary key is made, and the remaining attributes form a non-primary key in the logical schema.
             \item In case of $(1,N)$ cardinality on role $R_E$ (resp., $R_F$), the inclusion dependency $\sK_E \subseteq \sK_{RE}$ (resp., $\sK_F \subseteq \sK_{RF}$) holds in the schema, and the inclusions in the ontology hold in both directions.
             \item In case of $(1,1)$ cardinality on role $R_E$ (resp., $R_F$), the foreign key $\sK_E \rightarrow \sK_{RE}$ (resp., $\sK_F \rightarrow \sK_{RF}$) holds in the schema, and the first (resp., the second) inclusion axiom in the ontology holds in both directions.
             \end{compactitem}
         \end{minipage}
       } \\
       \midrule
    \multicolumn{4}{c}{\textbf{Schema Relationship with Identifier Alignment (SRa)}}\\
    \begin{minipage}{\erw}
      \includegraphics[scale=\erscale]{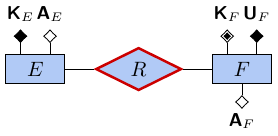}
    \end{minipage}
    &
    \begin{tabular}{l}
      \begin{tikzpicture}[dbpicture]
        \node (e)
          {$T_E(\pk{\sK\tikz\coordinate(kE);_E},\sA_E)$};
        \node[right=of e] (f)
          {$T_F(\pk{\sK_F},\sU\tikz\coordinate(uF);_F,\sA_F)$};
        \node[below=of e.south west,anchor=north west] (r)
          {$T_R(\pk{\sK\tikz\coordinate(kRE);_{RE},
              \sU\tikz\coordinate(kUF);_{RF}})$};
        \node[right=of r] (k) {$\key[T_F]{\sU_F}$};
        \draw[->] ($(kRE)+(0,\uoffset)$) -- ($(kE)-(0,\doffset)$);
        \draw[->] ($(kUF)+(0,\uoffset)$) |- ($(kUF)+(0,\uuoffset)$) -|
          ($(uF)-(0,\doffset)$);
        \end{tikzpicture}
    \end{tabular}
    &
    $\begin{array}{l@{~}l}
      s{:} & T_R \Join_{\sU_{RF}=\sU_F} T_F\\
      t{:} & p_R(\iri_{C_E}(\sK_{RE}),\iri_{C_F}(\sK_F))
    \end{array}$
    &
    $\begin{array}{l@{}}
      \exists p_R \sqsubseteq C_E \\
      \exists p_R^- \sqsubseteq C_F
     \end{array}$\\[5ex]
     \multicolumn{4}{l}{\small
       \begin{minipage}{1.48\textwidth}
         \setdefaultleftmargin{1em}{0em}{0em}{0em}{0em}{0em}
         Cardinality constraints are handled similarly as for \pat{SR}, with the difference that now the constraints now involve $U_{RF}$ and $U_F$.\\
         The case when both sets of attributes in $T_R$ require alignment is treated similarly.
       \end{minipage}
     } \\
    \midrule
    \multicolumn{4}{c}{\textbf{Schema Relationship with Merging (SRm)}}\\
    \begin{minipage}{\erw}
      \includegraphics[scale=\erscale]{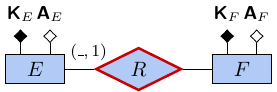}
    \end{minipage}
    &
    \begin{tabular}{l}
      \begin{tikzpicture}[dbpicture]
        \node (f)
          {$T_F(\pk{\sK\tikz\coordinate(kF);_F},\sA_F)$};
        \node[below=of f.south west,anchor=north west] (e)
          {$T_E(\pk{\sK_E},\sK\tikz\coordinate(kEF);_{EF},\sA_E)$};
        \draw[->] ($(kEF)+(0,\uoffset)$) |- ($(kEF)+(0,\uuoffset)$) -|
          ($(kF)-(0,\doffset)$);
      \end{tikzpicture}
    \end{tabular}
    & $\begin{array}{l@{~}l}
        s{:} & T_E\\
        t{:} & p_{EF}(\iri_{C_E}(\sK_E),\iri_{C_F}(\sK_{EF}))
      \end{array}$
    & $\begin{array}{l@{}}
        \exists p_{EF} \sqsubseteq C_E \\
        \exists p_{EF}^- \sqsubseteq C_F
      \end{array}$
      \\[5ex]
      \multicolumn{4}{l}{\small
       \begin{minipage}{1.48\linewidth}
         \setdefaultleftmargin{1em}{0em}{0em}{0em}{0em}{0em}
         Cardinality constraints are handled similarly as for \pat{SR}, with the catch that in case of $(0,1)$ cardinality on role $R_E$, then $\sK_{EF}$ is nullable. \\
         The alignment variant \pat{SRma/DRma}, where the foreign key references a non-primary identifier, is defined in the straightforward way.
       \end{minipage}
     } \\
         \midrule
    \multicolumn{4}{c}{\textbf{Schema Weak-Entity (SEw)}}\\
    \begin{minipage}{\erw}
      \includegraphics[scale=\erscale]{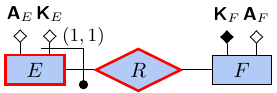}
    \end{minipage}
    &
    \begin{tabular}{l}
      \begin{tikzpicture}[dbpicture]
        \node (f)
          {$T_F(\pk{\sK\tikz\coordinate(kF);_F},\sA_F)$};
        \node[below=of f.south west,anchor=north west] (e)
          {$T_E(\pk{\sK_E,\sK\tikz\coordinate(kEF);_{EF}},\sA_E)$};
        \draw[->] ($(kEF)+(0,\uoffset)$) |- ($(kEF)+(0,\uuoffset)$) -|
          ($(kF)-(0,\doffset)$);
      \end{tikzpicture}
    \end{tabular}
    & $\begin{array}{l@{~}l}
        s{:} & T_E\\
        t{:} & C_E(\iri_{E}(\sK_E, \sK_{EF})), \\
             & \{d_A(\iri_E(\sK_E, \sK_{EF}),A)\}_{A \in \sK_E \cup \sA_E} \\
             & p_{EF}(\iri_{E}(\sK_E, \sK_{EF}),\iri_{C_F}(\sK_{EF}))
      \end{array}$
    & $\begin{array}{l@{}}
         \left\{\!\!\begin{array}{l}
                  \delta(d_A) \sqsubseteq C_E, \\
                  \rho(d_A) \sqsubseteq \mu(\tau(A)), \\
                  C_E \sqsubseteq \delta(d_{A})
                \end{array}\!\!
         \right\}_{A \in \sK_E\cup\sA_E} \\
        \exists p_{EF} \equiv C_E \\
        \exists p_{EF}^- \sqsubseteq C_F
      \end{array}$
      \\[5ex]
      \multicolumn{4}{l}{\small
        \begin{minipage}{1.48\textwidth}
          \setdefaultleftmargin{1em}{0em}{0em}{0em}{0em}{0em}
          Cardinality constraints are handled similarly as for \pat{SR}. \\
          Optional attributes are handled similarly as for \pat{SE}. \\
          The alignment variant \pat{SEwa/DEwa}, where the foreign key references a non-primary identifier, is defined in the straightforward way.
        \end{minipage}
      }\\
      \bottomrule
  \end{tabularx}
}
}
\caption{\small \label{tab:schema-driven-patterns-1}Schema-driven Patterns: Entities and Binary Relationships without Attributes.}
\end{table}


%% file: tables/schema-driven-2.tex
\begin{table}[t!]
  \centering
  {\linespread{0.96}\selectfont
  \resizebox{\textwidth}{!}{%
  \begin{tabularx}{1.57\textwidth}{l@{\hspace{2cm}}c@{\hspace{1pt}}l@{\hspace{0pt}}l@{}}
    \toprule
    \textsc{Conceptual Model} & \textsc{DB schema} & \textsc{Mappings} & \textsc{Ontology}\\
    \midrule
    \multicolumn{4}{c}{\textbf{Schema Reified Relationship (SRR)}}\\
    \begin{minipage}{\erw}
      \begin{tabular}{c}
        \includegraphics[scale=\erscale]{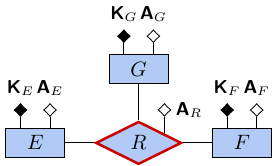} \\
        \bottomrule
        \includegraphics[scale=\erscale]{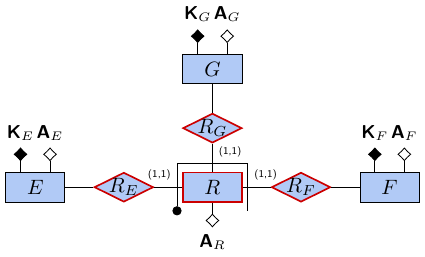} \\
      \end{tabular}
    \end{minipage}
    &
    \begin{tabular}{@{}l@{}}
      $\sK_R:=\sK_{RE}, \sK_{RF}, \sK_{RG}$\\
      \begin{tikzpicture}[dbpicture]
        \node (g)
          {$T_G(\pk{\sK\tikz\coordinate(kG);_G},\sA_G)$};
        \node[below=of g] (r)
          {$T_R(\pk{\sK\tikz\coordinate(kRE);_{RE},
            \sK\tikz\coordinate(kRF);_{RF},
            \sK\tikz\coordinate(kRG);_{RG}},
           \sA_R)$};
        \node[below=of r.south west,anchor=north west] (e)
          {$T_E(\pk{\sK\tikz\coordinate(kE);_E},\sA_E)$};
        \node[right=of e] (f)
          {$T_F(\pk{\sK\tikz\coordinate(kF);_F},\sA_F)$};
        \draw[->] ($(kRE)-(0,\doffset)$) -- ($(kE)+(0,\uoffset)$);
        \draw[->] ($(kRF)-(0,\doffset)$) |- ($(kRF)-(0,\ddoffset)$) -|
          ($(kF)+(0,\uoffset)$);
        \draw[->] ($(kRG)+(0,\uoffset)$) |- ($(kRG)+(0,\uuoffset)$) -|
          ($(kG)-(0,\doffset)$);
      \end{tikzpicture}
    \end{tabular}
    &
    $\begin{array}{l@{~}l}
      s{:} & T_R\\
      t{:} & C_R(\iri_R(\sK_R)), \\
           & \{d_A(\iri_R(\sK_R), A)\}_{A \in \sA_R}, \\
           & p_{RE}(\iri_R(\sK_R), \iri_{C_E}(\sK_{RE})), \\
           & p_{RF}(\iri_R(\sK_R), \iri_{C_F}(\sK_{RF})), \\
           & p_{RG}(\iri_R(\sK_R), \iri_{C_G}(\sK_{RG}))
    \end{array}$
    &
    $\begin{array}{l@{}}
      \exists p_{RE} \equiv C_R \\
      \exists p_{RE}^- \sqsubseteq C_E \\
      \exists p_{RF} \equiv C_R \\
      \exists p_{RF}^- \sqsubseteq C_F \\
      \exists p_{RG} \equiv C_R \\
      \exists p_{RG}^- \sqsubseteq C_G \\
      \left\{
       \begin{array}{l}
         \delta(d_A) \sqsubseteq C_R, \\
         \rho(d_A) \sqsubseteq \mu(\tau(A)), \\
         C_R \sqsubseteq \delta(d_{A})
       \end{array}
       \right\}_{A \in \sA_R}
     \end{array}$\\[10ex]
     \multicolumn{4}{l}{\small
       \begin{minipage}{1.54\linewidth}
         \setdefaultleftmargin{1em}{0em}{0em}{0em}{0em}{0em}
         \pat{SRR} applies whenever there are three or more participating roles, or when the relationship has attributes. Given the nature of RDF graphs, in order to handle these cases we need reification, hence this pattern requires a change in the conceptual model (see ER-diagram below the line). After reification, we apply the patterns discussed for binary relationships (cardinality constraints, weak entities, and optional attributes are handled as discussed). Observe that, in the conversion to \owlql, the identification constraint on $R$ is lost (similarly to other identifiers). 
       \end{minipage}
   } \\
     \midrule
    \multicolumn{4}{c}{\textbf{Schema Hierarchy (SH)}}\\
    \begin{tabular}{c}
      \begin{minipage}{\erw}
        \centering
        \includegraphics[scale=\erscale]{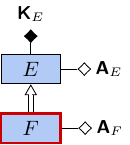}
      \end{minipage}
    \end{tabular}
    &
    \begin{tabular}{l}
      \begin{tikzpicture}[dbpicture]
        \node (e)
          {$T_E(\pk{\sK\tikz\coordinate(kE);_E},\sA_E)$};
        \node[below=of e.south west,anchor=north west] (f)
          {$T_F(\pk{\sK\tikz\coordinate(kF);_{FE}},\sA_F)$};
        \draw[->] ($(kF)+(0,\uoffset)$) -- ($(kE)-(0,\doffset)$);
      \end{tikzpicture}
    \end{tabular}
    &
      $\begin{array}{l@{~}l}
        s{:} & T_F \\
        t{:} & C_F(\iri_{C_E}(\sK_{FE})), \\
             & \{d_A(\iri_{C_E}(\sK_{FE}),A)\}_{A \in \sA_F}
       \end{array}$
    &
      $\begin{array}{l@{}}
         C_F \sqsubseteq C_E \\
         \left\{
         \begin{array}{l}
           \delta(d_A) \sqsubseteq C_F, \\
           \rho(d_A) \sqsubseteq \mu(\tau(A)), \\
           C_F \sqsubseteq \delta(d_{A})
         \end{array}
         \right\}_{A \in \sA_F}
      \end{array}$
      \\[5ex]
     \multicolumn{4}{l}{\small
       \begin{minipage}{1.54\linewidth}
         \setdefaultleftmargin{1em}{0em}{0em}{0em}{0em}{0em}
         Optional attributes are handled as in \pat{SE}.
       \end{minipage}
   } \\
    \midrule
    \multicolumn{4}{c}{\textbf{Schema Hierarchy with Identifier Alignment (SHa)}}\\
    \begin{tabular}{c}
      \begin{minipage}{\erw}
        \centering
        \includegraphics[scale=\erscale]{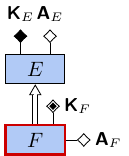}
      \end{minipage}\\
      \bottomrule
      \begin{minipage}{\erw}
        \centering
        \includegraphics[scale=\erscale]{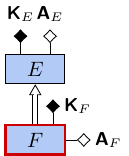}
      \end{minipage}
    \end{tabular}
    &
    \begin{tabular}{@{}l@{}}
      \begin{tikzpicture}[dbpicture]
        \node (e)
          {$T_E(\pk{\sK\tikz\coordinate(kE);_E},\sA_E)$};
        \node[below=of e.south west,anchor=north west] (f)
          {$T_F(\pk{\sK\tikz\coordinate(kF);_F},\sK\tikz\coordinate(uF);_{FE},
           \sA_F)$};
        \node[below=of e.east,anchor=west] (k)
          {$\key[T_F]{\sK_{FE}}$};
        \draw[->] ($(uF)+(0,\uoffset)$) |- ($(uF)+(0,\uuoffset)$) -|
          ($(kE)-(0,\doffset)$);
        \end{tikzpicture} \\
      \bottomrule
      \begin{tikzpicture}[dbpicture]
        \node (e)
          {$T_E(\pk{\sK\tikz\coordinate(kE);_E},\sA_E)$};
        \node[below=of e.south west,anchor=north west] (f)
          {$V_F(\sK\tikz\coordinate(kF);_F,\pk{\sK\tikz\coordinate(uF);_{FE}},
            \sA_F) = T_F$ };
        \node[right=of e.east,anchor=west] (k)
          {$\key[V_F]{\sK_F}$ };
        \draw[->] ($(uF)+(0,\uoffset)$) |- ($(uF)+(0,\uuoffset)$) -|
          ($(kE)-(0,\doffset)$);
      \end{tikzpicture}
    \end{tabular}
    &
      $\begin{array}{l@{~}l}
         s{:} & V_F \\
         t{:} & C_F(\iri_{C_E}(\sK_{FE})), \\[0.5ex]
         & \{d_A(\iri_{C_E}(\sK_{FE}),A)\}_{A \in \sK_F \cup \sA_F}
      \end{array}$
    &
      $\begin{array}{l@{}}
        C_F \sqsubseteq C_E \\
         \left\{
         \begin{array}{l}
           \delta(d_A) \sqsubseteq C_F, \\
           \rho(d_A) \sqsubseteq \mu(\tau(A)), \\
           C_F \sqsubseteq \delta(d_{A})
         \end{array}
           \right\}_{A \in \sK_F\cup\sA_F}
    \end{array}$
    \\[10ex]
    \multicolumn{4}{l}{\small
      \begin{minipage}{1.54\linewidth}
         \setdefaultleftmargin{1em}{0em}{0em}{0em}{0em}{0em}
        In this pattern, the ``alignment'' is meant to align the primary identifier used in the child entity to the primary identifier used in the parent entity. The other two possiblities for the application of the pattern are:

        \begin{compactitem}
        \item the foreign key in the child entity is the primary key of that entity, and references a non-primary key of the parent entity;
        \item the foreign key in the child entity is a non-primary key of that entity, and  references a non-primary key of the parent entity.
        \end{compactitem}
        We here depict the most common scenario, where the foreign key points to the primary key of the parent entity.\\
        Observe that this pattern requires a change in the conceptual model (essentially keeping track the attributes used for identifying the objects of the subclass). \\
        Optional attributes are handled as in \pat{SH}.
      \end{minipage}
    }\\
    \bottomrule
  \end{tabularx}
}
}
\caption{\label{tab:schema-driven-patterns-2}\small Schema-driven Patterns: Relationships with Attributes, N-ary Relationships, and Hierarchies. For patterns yielding views, we show the views together with the DB schema, separating them from the original tables using a thick horizontal bar. We use a similar notation for changes in the conceptual model.}
\end{table}


%% file: 3.3-mapping-data.tex
\subsection{Data Driven Mapping Patterns}
\label{sec:mapping-data}

\input{tables/data-driven.tex}

Data-driven patterns are mapping patterns that depends both on the schema and on the actual data in the DB. They are not limited to the variants corresponding to the schema-driven patterns, but they also comprehend specific patterns that do not have a corresponding schema version, e.g., due to \emph{denormalized tables}. Such patterns, for which we provide a detailed description below, are shown in Table~\ref{tab:data-driven-patterns}. Similarly as we did for schema-driven patterns, we provide an example and references to related literature for each data-driven pattern.

\mypar{Data Entity with Merged 1-N Relationship and Entity (\pat{DR1Nm})} This pattern describes the situation where both the relationship and the participating entity have been merged into a table. It considers a table $T_E$ that has, besides its primary key $\sK_E$, also attributes $\sK_F$ which functionally determine attributes $\sA_F$. Observe that the latter condition is not possible if the DB schema is in 3rd\footnote{A straightforward variant of this pattern, violating the 2nd normal form, could be added to our list.} normal form~\cite{Codd71}.
When this pattern is applied, the key $\sK_F$ and the attributes $\sA_F$ that
go along with it, can be projected out from $T_F$, resulting in a view $V_F$
to which further patterns can be applied, for instance \pat{SE}.
An additional view $V_E$ is also created, representing the entity $E$.
\\
\textsl{Example}: A table \ex{restaurant} containing information about restaurants, their unique supplier (identified by a code), and the address of the supplier. The
supplier identifier, which is not a key for \ex{restaurant}, uniquely determines the address of the supplier. Table \ex{restaurant} will be vertically partitioned into two views, \ex{restaurant} and \ex{suppliers}, that will later be linked to their respective ontology classes \ex{Restaurant} and a \ex{Supplier} through the applications of \pat{SE} on the fresh views. An object property is created
connecting restaurants to their suppliers.

~\\
\noindent \resizebox{\textwidth}{!}{%
\begin{tabular}{c@{\hspace{2pt}}c}
  \toprule
  \textsc{Conceptual Model} & \textsc{DB schema}\\
  \multicolumn{2}{c}{(Datatypes Omitted)} \\
  \midrule
  \begin{minipage}{\erw}
    \centerline{\qquad \includegraphics[scale=1]{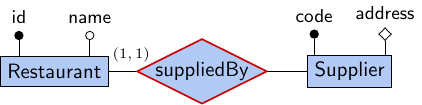}}
  \end{minipage}
  &
    \begin{tabular}{l}
      \begin{tikzpicture}[dbpicture]
        \node (te) {\normalsize restaurant(\pk{id}, name, supC, supA)};
        \node[below=of te.south west, anchor=north west] (fd) {\normalsize $\fd[{\sf restaurant}]{\sf supC}{\sf code}$};
      \end{tikzpicture}
    \end{tabular}
  \\[2em]
  \midrule
  \textsc{Mappings} & \textsc{Ontology} \\
  \midrule
  \begin{tabular}{l@{~}l}
    \ex{s{:}} & \ex{SELECT id, supC FROM V-restaurant} \\
    \ex{t{:}} & \ex{:restaurant/\{id\} :suppliedBy :supplier/\{supC\}} . \\
    \multicolumn{2}{l}{V-restaurant := SELECT id, name, supC FROM restaurant} \\
    \end{tabular}
  &
    \begin{tabular}{ll}
      :suppliedBy a owl:ObjectProperty . \\
      :suppliedBy rdfs:domain :Restaurant . & :suppliedBy rdfs:range :Restaurant . \\
      :Restaurant rdfs:subClassOf _:r1 . & _:r1 a owl:Restriction . \\
      _:r1 owl:onProperty :suppliedBy . & _:r1 owl:someValuesFrom rdfs:Literal . \\
    \end{tabular} \\
  \bottomrule
\end{tabular}
}

~\\
\textsl{References}: Slight variants for this pattern can be found in the literature:
\begin{compactitem}
\item BootOX~\cite{JKZH*15} reports a mappings generation strategy for situations similar to the one of Pattern \pat{DR1Nm}. However, details of how this is actually carried out are not provided.
\item Pattern \emph{Relationship: One to Many with Duplicates} from~\cite{sequeda-book} handles the same situation of \pat{DR1Nm}, proposing a similar solution. Such work also proposes an alternative solution in Pattern \emph{Relationship: One to Many without Duplicates}, where the primary identifier of table $T_E$ is used to create both URIs for $C_E$ and $C_F$.
\end{compactitem}

\mypar{Data Entity with Optional Participation in a Relationship (DH01)}%
This pattern describes the situation where non-mandatory relationship is transformed into a mandatory one through the introduction of a subconcept on one of its participating roles. It is characterized by a table $T_E$ that represents the merge of child entity $E_R$ into a father entity $E$, and $E_R$ has a mandatory participation in a relationship $R$. The join between the table $T_R$ and $T_E$ identifies the objects in $E$ instances of $E_R$, and is used in a mapping to create instances of the concept $C_{R_E}$, as well as the object property $R$ connecting $E_R$ to $F$. This pattern produces a view $V_{E_R}$, to which further patterns can be applied.
\\
\textsl{Example}: A students table and a table connecting students to undergraduate courses. Each student participating such relationship is an undergraduate student.
\begin{center}
\noindent \resizebox{.9\textwidth}{!}{%
\begin{tabular}{c@{\hspace{2pt}}c}
  \toprule
  \textsc{Conceptual Model} & \textsc{DB schema}\\
  \multicolumn{2}{c}{(Datatypes Omitted)} \\
  \midrule
  \begin{minipage}{\erw}
    \centerline{\qquad \includegraphics[scale=1]{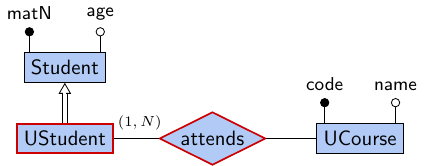}}
  \end{minipage}
  &
    \begin{tabular}{l}
      \begin{tikzpicture}[dbpicture]
        \node (e)
        {\normalsize student(\pk{ma\tikz\coordinate(kE);tN}, age)};
        \node[right=of e] (f)
        {\normalsize ucourse(\pk{co\tikz\coordinate(kF);de},name)};
        \node[below=of e.south west,anchor=north west] (r)
        {\normalsize attends(\pk{sm\tikz\coordinate(kRE);atN,
            cc\tikz\coordinate(kRF);ode})};
        \draw[->] ($(kRE)+(0.3mm,\uoffset)$) -| ($(kRE)+(0.3mm,\uuoffset)$) -| ($(kE)-(0,\doffset)$);
        \draw[->] ($(kRF)+(0,\uoffset)$) |- ($(kRF)+(0,\uuoffset)$) -|
        ($(kF)-(0,\doffset)$);
      \end{tikzpicture}\\
    \end{tabular}
  \\[2em]
  \midrule
  \textsc{Mappings} & \textsc{Ontology} \\
  \midrule
  \begin{tabular}{l@{~}l}
    \ex{s1{:}} & \ex{SELECT smatN FROM v-UStudent} \\
    \ex{t1{:}} & \ex{:student/\{smatN\} a :UStudent} . \\
    \ex{s2{:}} & \ex{SELECT smatN, ccode FROM attends} \\
    \ex{t2{:}} & \ex{:student/\{smatN\} :attends :course/\{ccode\}} . \\
    \multicolumn{2}{l}{v-UStudent := SELECT smatN FROM attends}
    \end{tabular}
  &
    \begin{tabular}{ll}
      :UStudent a owl:Class . & :attends a owl:ObjectProperty . \\
      :attends rdfs:domain :UStudent . & :attends rdfs:range :UCourse . \\
      :UStudent rdfs:subClassOf _:r1 . & _:r1 a owl:Restriction . \\
      _:r1 owl:onProperty :attends . & _:r1 owl:someValuesFrom owl:Thing . \\
    \end{tabular} \\
  \bottomrule
\end{tabular}
}
\end{center}
\textsl{References}: To the best of our knowledge, only BootOX~\cite{JKZH*15} reports a mappings generation strategy handling a scenario that resembles the one of Pattern \pat{DH01}. However, details of how this is actually carried out are not provided.

\mypar{Clustering Entity to Concept/Data Property/Object Property \pat{(CE2C/CE2D/CE2O)}}\\
Such patterns are characterized by an entity $E$ and a \emph{derivation rule} defining sub-entities of $E$ according to the values for attributes $\sB$ in $E$. Instances in these sub-entities can be mapped to objects in the subclasses $C_E^{\mathfrak{p}}$ of the ontology (\pat{CE2C}), to objects connected through a data property to some literal constructed through a \emph{value invention} function $\xi$ applied on a partition $\mathfrak{p}$ (\pat{CE2D}), or to objects (i.e., IRIs) constructed through an \emph{object invention} function $\gamma$ applied on $\mathfrak{p}$ (\pat{CE2O}).
\\
\textsl{Example}: A table contains people with an attribute defining their
gender and ranging over '\texttt{M}' or '\texttt{F}'. The ontology defines a
data property \ex{hasGender}, ranging over the two RDF literals \texttt{"Male"}
and \texttt{"Female"}. Then, pattern \pat{CE2D} clusters the table according to
the gender attribute, so as to obtain objects to be linked to either of the two
RDF literals.
\begin{center}
\noindent \resizebox{.9\textwidth}{!}{%
\begin{tabular}{c@{\hspace{2pt}}c}
  \toprule
  \textsc{Conceptual Model} & \textsc{DB schema}\\
  \multicolumn{2}{c}{(Datatypes Omitted)} \\
  \midrule
  \begin{minipage}{\erw}
    \begin{tabular}{c}
      \centerline{\includegraphics[scale=1]{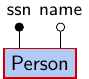}}
      \\
      \begin{tabular}{c}
        \sf sex = 'M' $\rightarrow$ Male \\
        \sf sex = 'F' $\rightarrow$ Female \\
      \end{tabular}
    \end{tabular}
  \end{minipage}
  &
    \begin{tabular}{l}
      \begin{tikzpicture}[dbpicture]
        \node (e) {\normalsize person(\pk{ssn}, name, sex)};
      \end{tikzpicture}\\
    \end{tabular}
  \\[2em]
  \midrule
  \textsc{Mappings} & \textsc{Ontology} \\
  \midrule
  \begin{tabular}{l@{~}l}
    \ex{s1{:}} & \ex{SELECT ssn FROM person WHERE sex='M'} \\
    \ex{t1{:}} & \ex{:person/\{ssn\} a :Male} . \\
    \ex{s2{:}} & \ex{SELECT ssn FROM person WHERE sex='F'} \\
    \ex{t2{:}} & \ex{:person/\{ssn\} a :Female} .
    \end{tabular}
  &
    \begin{tabular}{ll}
      :Male a owl:Class . & :Female a owl:Class . \\
      :Male owl:subClassOf :Person & :Female owl:subClassOf :Person .
    \end{tabular} \\
  \bottomrule
\end{tabular}
}
\end{center}
\textsl{References}: For what concerns the \pat{CE2C} variant, the clustering pattern is related to a number of works:
\begin{compactitem}
\item In BootOX~\cite{JKZH*15}, an automated approach that mentiones ``clustering'' is reported. However, the clusters in their approach are sets of ``similar'' tuples. This is different from our Pattern~\pat{CE2C}, which instead requires a set of columns whose values explicitly determine the different clusters.
\item In~\cite{sequeda-book}, Patterns \emph{Complex Concept: Conditions} and \emph{Complex Concept: Data as Concept} share the idea of imposing a condition in order to identify subsets of a table and creating concepts out of them. However, it has to be noted that ontology axioms are not in the scope of that work. Pattern \emph{Complex Concept Attribute: Constant Value} is similar to our Pattern~\pat{CE2D}, since it associates objects satisfying a certain equality filter condition to a constant data value.
\end{compactitem}

\begin{example}
  Consider again the conceptual model, DB schema, mappings, and ontology derived in Example~\ref{e:example-schema-driven}. Assume a derivation rule on entity \ex{Person} identifying two sub-entities: an entity representing those whose age is greater or equal than 18, which we call \ex{OfAge}, and another one representing the others, which we call \ex{UnderAge}. Under these assumptions, we can apply pattern \pat{CE2C} and obtain:
  ~\\[1ex]
  \noindent \resizebox{\textwidth}{!}{%
    \begin{tabular}{c@{\hspace{2pt}}c}
      \toprule
      \textsc{Mappings} & \textsc{Ontology} \\
      \multicolumn{2}{c}{(Datatypes and Mandatory Participations Omitted)} \\
      \midrule
      \sf
      \begin{tabular}{l@{~}l}
        \ex{s1{:}} & \ex{SELECT ssn FROM person where age >= 18} \\
        \ex{t1{:}} & \ex{:Person/ssn=\{ssn\}} a :OfAge . \\
        \ex{s2{:}} & \ex{SELECT ssn FROM person where age < 18} \\
        \ex{t2{:}} & \ex{:Person/ssn=\{ssn\}} a :Underage . \\
      \end{tabular}
                        & \sf
                          \begin{tabular}{ll}
                            :OfAge a owl:Class . & :UnderAge a owl:Class . \\
                            :OfAge owl:subClassOf :Person . & :UnderAge owl:subClassOf :Person .
                          \end{tabular} \\
      \bottomrule
    \end{tabular}
}
\end{example}


%% file: tables/data-driven.tex
\begin{table}[t!]
  \centering
  {\linespread{0.96}\selectfont
    \resizebox{\textwidth}{!}{%
      \begin{tabularx}{1.5\textwidth}{p{3.5cm}cp{4.7cm}p{4cm}}
        \toprule
        \textsc{Conceptual Model} & \textsc{DB schema} & \textsc{Mappings} &
        \textsc{Ontology} \\
        \midrule
        \multicolumn{4}{c}{\textbf{Data Entity with Merged 1-N Relationship and
            Entity (DR1Nm)}}\\
        \begin{minipage}{\erw}
          \includegraphics[scale=\erscale]{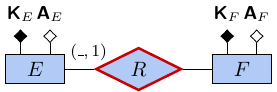}
        \end{minipage}
        &
        \begin{tabular}{@{}l@{}}
          \begin{tikzpicture}[dbpicture]
            \node (te) {$T_E(\pk{\sK_E},\sA_E,\sK_F,\sA_F)$};
            \node[below=of te.south west, anchor=north west] (fd) {$\fd[T_E]{\sK_F}{\sA_F}$};
          \end{tikzpicture}\\
          \bottomrule
          \begin{tikzpicture}[dbpicture]
            \node (ve)
            {$V_E(\pk{\sK_E}, \sK\tikz\coordinate(kEF);_{EF}, \sA_E) = \pi_{\sK_E, \sK_F, \sA_E} (T_E)$};
            \node[below=of ve.south west,anchor=north west] (vf)
            {$V_F(\pk{\sK\tikz\coordinate(kF);_F},\sA_F) = \pi_{\sK_F,\sA_F} (T_E)$};
            \draw[->] ($(kEF)-(0,\doffset)$) -- ($(kEF)-(0,\ddoffset)$) -| ($(kF)+(0,\uoffset)$);

          \end{tikzpicture}
        \end{tabular}
        &
        $\begin{array}{l@{~}l}
           s{:} & V_E \\
           t{:} & p_R(\iri_E(\sK_E),\iri_F(\sK_{EF}))
         \end{array}$
         &
         $\begin{array}{l@{}}
            \exists p_R \sqsubseteq C_E \\
            \exists p_R^- \sqsubseteq C_F
          \end{array}$
          \\[4em]
          \multicolumn{4}{l}{\footnotesize
            \begin{minipage}{1.46\textwidth}
              Mappings and ontology axioms for classes $C_E$ and $C_F$, not shown here, conform to Pattern~\pat{SE/DE} on the newly introduced views $V_E$ and $V_F$.\\
              All considerations on cardinality constraints and optional attributes described for \pat{SRm} extend to this pattern in the natural way.
            \end{minipage}
          }\\
          \midrule
          \multicolumn{4}{c}{\textbf{Data Entity with Optional Participation in a Relationship (DH01)}}\\
          \begin{minipage}{\erw}
            \begin{tabular}{l}
              \includegraphics[scale=\erscale]{img/er-SR.pdf} \\[.5em]
              \bottomrule
              \includegraphics[scale=\erscale]{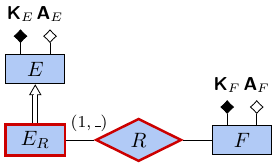}
            \end{tabular}
          \end{minipage}
          &
          \begin{minipage}{.38\linewidth}
            \begin{tabular}{@{}l@{}}
              \begin{tikzpicture}[dbpicture]
                \node (e)
                {$T_E(\pk{\sK\tikz\coordinate(kE);_E},\sA_E)$};
                \node[right=of e] (f)
                {$T_F(\pk{\sK\tikz\coordinate(kF);_F},\sA_F)$};
                \node[below=of e.south west,anchor=north west] (r)
                {$T_R(\pk{\sK\tikz\coordinate(kRE);_{RE},
                    \sK\tikz\coordinate(kRF);_{RF}})$};
                \draw[->] ($(kRE)+(0.3mm,\uoffset)$) -- ($(kE)-(0,\doffset)$);
                \draw[->] ($(kRF)+(0,\uoffset)$) |- ($(kRF)+(0,\uuoffset)$) -|
                ($(kF)-(0,\doffset)$);
              \end{tikzpicture}\\
              \bottomrule
              \begin{tikzpicture}[dbpicture]
                \node (e) at (0,0)
                {$T_E(\pk{\sK\tikz\coordinate(kE);_E},\sA_E)$};
                \node (f) at (2,0)
                {$T_F(\pk{\sK\tikz\coordinate(kF);_F},\sA_F)$};
                \node (r) at (1,-.5)
                {$T_R(\pk{\sK\tikz\coordinate(kRE);_{RE},
                    \sK\tikz\coordinate(kRF);_{RF}})$};
                \node[below=of r] (Er) {
                  $\begin{array}{l}
                     V_{E_R}(\pk{\sK\tikz\coordinate(vKE);_E}) ={} \pi_{\sK_{RE}} (T_R)
                   \end{array}$
                 };
                 \draw[<->] ($(kRE)+(0.3mm,-1mm)$) |- ($(kRE)-(0,\ddoffset)$) -| ($(vKE)-(-1mm,-\uoffset)$);
                 \draw[->] ($(kRF)+(0,\uoffset)$) |- ($(kRF)+(0,\ddoffset)$) -| ($(kF)-(0,1mm)$);
                 \draw[->] ($(vKE) + (-1mm,\uoffset)$) |- ($(vKE) + (-5mm,\uuoffset)$) -| ($(kE) + (0,-\doffset)$);
               \end{tikzpicture}
            \end{tabular}
          \end{minipage}
          &
          $\begin{array}{l@{~}l}
             s_1{:} & V_{ER} \\
             t_1{:} & C_{E_R}(\iri_{C_E}(\sK_{E})) \\
             s_2{:} & T_R \\
             t_2{:} &  p_R(\iri_{C_E}(\sK_{RE}),\iri_{C_F}(\sK_{RF}))
           \end{array}$
           &
           $\begin{array}{l@{}}
              C_{E_R} \sqsubseteq C_E \\
              \exists p_R \equiv C_{E_R} \\
              \exists p_R^- \sqsubseteq C_F
            \end{array}$
            \\[4em]
            \multicolumn{4}{l}{\footnotesize
              \begin{minipage}{1.46\textwidth}
                This pattern extends in the natural way to the variants with identifier alignment (see pattern \pat{SRa}) and reified relationship (see pattern \pat{SRR}).\\
                All considerations on cardinality constraints and optional attributes described for \pat{SE} and \pat{SRm} extend to this pattern in the natural way.
              \end{minipage}
            }
            \\
            \midrule
            \multicolumn{4}{c}{\textbf{Clustering Entity to Class (CE2C)}}\\
            \begin{minipage}{\erw}
                \begin{tabular}{c}
                \includegraphics[scale=\erscale]{img/er-SE.pdf} \\
                $\begin{array}{c}
                   \sB \subseteq \sK\cup\sA\\
                   \partition(\sB,E)
                 \end{array}
                 $ \\
                  partitions to classes
                \end{tabular}
             \end{minipage}
             & $\begin{array}{l}
                  T_E(\pk{\sK}, \sA) \\
                  \sB \subseteq \sK\cup\sA \\
                  \partition(\sB,E)
                \end{array}$
                & $\left\{
                  \begin{array}{l@{~}l}
                    s: & \sigma_{\mathfrak{p}}(T_E) \\
                    t: & C_E^{\mathfrak{p}}(\iri_{C_E}(\sK))
                  \end{array}
                \right\}_{\mathfrak{p} \in \partition(\sB,E)}$
                & $\begin{array}{l}
                     \{C_E^{\mathfrak{p}} \sqsubseteq C_E \}_{\mathfrak{p} \in \partition(\sB,E)}
                   \end{array}$\\
                   \midrule
                   \multicolumn{4}{c}{\textbf{Clustering Entity to Object (CE2O)}}\\
                   \begin{minipage}{\erw}
                     \begin{tabular}{c}
                       \includegraphics[scale=\erscale]{img/er-SE.pdf}\\
                       $\begin{array}{c}
                          \sB \subseteq \sK\cup\sA\\
                          \partition(\sB,E)
                        \end{array}
                        $\\
                        partitions to objects
                     \end{tabular}
                    \end{minipage}
                    & $\begin{array}{l}
                         T_E(\pk{\sK}, \sA) \\
                         \sB \subseteq \sK\cup\sA \\
                         \partition(\sB,E)
                       \end{array}$
                       & $
                       \left\{
                         \begin{array}{l@{~}l}
                           s: & \sigma_{\mathfrak{p}}(T_E) \\
                           t: & p_{\sB}(\iri_{C_E}(\sK), \gamma_{\mathfrak{p}})
                         \end{array}
                       \right\}_{\mathfrak{p} \in \partition(\sB,E)}$
                       & \hspace{.7em}$\exists p_{\sB} \sqsubseteq C_E$ \\
                       \midrule
                       \multicolumn{4}{c}{\textbf{Clustering Entity to Data Value (CE2D)}}\\
                       \begin{minipage}{\erw}
                         \begin{tabular}{c}
                           \includegraphics[scale=\erscale]{img/er-SE.pdf}\\
                           $\begin{array}{c}
                              \sB \subseteq \sK\cup\sA\\
                              \partition(\sB,E)
                            \end{array}
                            $\\
                            partitions to values
                         \end{tabular}
                        \end{minipage}
                        & $\begin{array}{l}
                             T_E(\pk{\sK}, \sA) \\
                             \sB \subseteq \sK\cup\sA \\
                             \partition(\sB,E)
                           \end{array}$
                           &
                           $
                           \left\{
                             \begin{array}{l@{~}l}
                               s: & \sigma_{\mathfrak{p}}(T_E) \\
                               t: & d_{\sB}(\iri_{C_E}(\sK), \xi_{\mathfrak{p}})
                             \end{array}
                           \right\}_{\mathfrak{p} \in \partition(\sB,E)}$
                           &
                           \hspace{.7em}$\delta(d_{\sB}) \sqsubseteq C_E$
                           \\
                           \bottomrule
                         \end{tabularx}
                       }
                     }
                     \caption{\label{tab:data-driven-patterns}\small Data-driven Patterns}
                   \end{table}


%% file: 3.4-variations.tex
\subsection{Variations and Combinations}

More complex patterns arise from the combination of the patterns described so far. For instance, recall the example we discussed for pattern \pat{DH01}. Graduate students, which are a by-product of the application of such pattern, might be in relationship with an entity \ex{Graduation}. The object property capturing the relationship might be created by applying pattern \pat{DR}. In our analysis, we have observed that combinations are quite common in those \vkg specifications where the DB has been created independently from the ontology.

Another important variation is the one introduced by modifiers, such as \emph{value invention or combination}, in which DB values are used and combined to get RDF literals, typically by relying on R2RML \emph{templates}. We have already encountered an instance of value invention, specifically when we introduced the \pat{CE2D} pattern.


%% file: 3.5-automatic-discovery.tex
\subsection{Automatic Discovery of Data-driven Patterns}

When it comes to discovering data-driven patterns, our methodology may benefit from techniques that were developed in the research discipline of schema matching~\cite{rahm2001survey}. Over the years, the proposed methods were shown to serve as a solid basis to handle small-scale schemata, typically encountered as a part of a mapping process~\cite{shragaadnev}.
Schema matching becomes handy when a pattern involves two (or more) under-specified schemata. By way of motivation, consider the case of an implicit relationship (pattern \pat{DR}). In such a case, the mapping may consider several relation pair candidates that may be semantically interpreted as representing a missing relationship. Let $T_E$ and $T_F$ with primary keys $\sK_E = K_{E_1}, \ldots, K_{E_n}$ and $\sK_F = K_{F_1}, \ldots, K_{F_m}$, respectively, be a relation candidate pair. A matching process between $\sK_E$ and $\sK_F$ aligns their attributes using \emph{matchers} that utilize matching cues such as attribute names, instance data, schema structure, etc. Accordingly, the matching process yields similarity values (typically a real number in $[0,1]$) between $K_{E_i} \in \sK_E$ and $K_{F_j} \in \sK_F$. These values are then used to deduce a \emph{match} $\sigma(\sK_E, \sK_F)$. Such a match may comply with different constraints as set by the environment, e.g., a one-to-one matching. For example, the similarity values may be assigned using a string similarity matcher like~\cite{gal2011uncertain}
{\small
\[\frac{len(K_{E_i}.name\cap K_{F_j}.name)}{max(len(K_{E_i}.name), len(K_{F_j}.name))}
\]
}
and a match may be inferred using a threshold selection rule~\cite{do2002coma}. Once a match is obtained, it may serve as a realization of a schema relationship mapping pattern. Obviously, not all tables have relationships. Thus, one should decide which of the generated relationships should be included in the final mapping. To do so, we should assess the quality of the matching outcome, allowing to rank among them~\cite{gal2019learning} (e.g., selecting matches with high similarity values to be included in the final mapping). Match quality may also be learned as a domain specific input introducing other quality measures to the usefulness of a match.

The schema matching literature offers a mechanism to map multiple schemata to a global schema, which bears similarity to the \pat{SR1Nm} pattern. \emph{Schema cover}~\cite{gal2013completeness} matches parts of schemata (subschemata) with concepts, using schema matching techniques, aiming at covering all attributes of the global schema with minimum number of overlaps between the subschemata. Using a similar approach, one can ``cover'' a schema by using multiple ontology concepts to generate an \pat{SR1Nm} pattern. Recalling the example we discussed for \pat{SR1Nm}, using the schema cover methodology, we can cover the \ex{restaurants} table using the properties relative to restaurants defined in the ontology, and the data property \ex{address} from the concept \ex{Supplier}.


%% file: 5-pattern-usage.tex
\section{Usage Scenarios for \vkg Patterns}

\label{sec:discussion}
We now comment on how having a catalog of patterns for \vkg specifications is instrumental in a number of usage scenarios. The provided list is by no means complete, but gives a fair account on the usefulness of our approach.

\mypar{Debugging of a \vkg Specification} This scenario arises when a full \vkg specification is already in place and must be debugged. If the conceptual model is available, for instance in the form of knowledge by the data curators, each component of the specification can be checked for compliance against the patterns. Such checks are more effective on ontologies that are closer to the actual conceptual representation of the DB, and can answer questions such as whether the relationships are being correctly represented in terms of their domains, range, and mandatory participations, or whether subclasses are defined using the correct templates. Ontologies whose structure is far from the DB ontology require more effort. A possibility is to adopt a two-step process of deriving the DB ontology first, conforming to the patterns, and then checking whether there exists a ``lossless'' \emph{alignment} between the DB ontology and the target one (in line with the intution of Figure~\ref{f:db-conc-onto}).

\mypar{Conceptual Model Reverse Engineering} Another relevant scenario arising when a full \vkg specification is given, is that of inferring a conceptual model of the DB that represents the domain of interest by reflecting the content of the \vkg specification. Here the ontology provides the main source to reconstruct entities, attributes, and relationships, while the DB and the mappings provide the basis to ground the conceptual model in the actual DB, and to infer additional constraints that are not captured by the ontology (e.g., for limited expressivity of \owlql). As for the debugging case, also this approach works best if the ontology is semantically close to the conceptual model.

\mypar{Mapping Bootstrapping} In this scenario, the DB and the ontology are given, but mappings relating them are not. We envision this as a two-step process: in the first step, we use our patterns to derive a well-structured DB ontology; in the second step, we rely on techniques coming from the field of \emph{ontology matching or alignment}~\cite{survey-onto-matching} to produce so-called \emph{alignment mappings} which relate the DB ontology to the actual target ontology. Schema patterns are the most suitable ones to automatically guide the bootstrapping process. When patterns contain tables that merge multiple entities/relationships, the presence of a conceptual model becomes crucial to disambiguate the mappings to be bootstrapped. This is, e.g., the case for \pat{DR1Nm} and the patterns based on clustering.
If the conceptual model is not available in this tricky case, boostrapping can still be attempted by relying on schema matching techniques~\cite{rahm2001survey}, as done in BootOX~\cite{JKZH*15}.
Specifically, schema matching comes handy when a pattern involves two (or more) under-specified schemata. For instance, in the case of pattern \pat{DR}, \emph{pair candidates} between primary keys can be \emph{matched} in order to make implicit relationships explicit. This can be done through matchers (such as string similarity matchers ~\cite{gal2011uncertain}) that employ attribute names, instance data, schema structure, etc. To separate genuine relationships from false positives generated by poor matchers, ranking techniques have to be employed~\cite{gal2019learning}.

\mypar{Ontology+Mapping Bootstrapping} Here, neither the ontology nor the mappings are given as input, and have to be synthesized. This scenario can be tackled as the \textbf{Mapping Bootstrapping} one, by omitting the second step. As already discussed, best results are to be expected when a conceptual model is available, since the obtained ontology will likely be closer to the level of abstraction expected by the domain experts.

\mypar{\vkg Bootstrapping} In this scenario, we just have a conceptual model of the domain, and the goal is to set up a \vkg specification. 
The conceptual model can be then transformed into a  normalized DB schema using well-established \emph{relational mapping} techniques (e.g.,~\cite{HaMo10}). At the same time, as pointed above, a direct encoding into ontology axioms can be applied to bootstrap the ontology. The generation of mappings becomes then a quite trivial task, considering that the induced DB and ontology are very close in terms of abstraction.
This setting resembles, in spirit, that of \emph{object-relational mapping}, used in software engineering to instrument a DB and corresponding access mechanisms starting from classes written in object-oriented code. 


%% file: 6-scenarios.tex
\section{Analysis of Scenarios}
\label{sec:scenarios}

In this section we look at a number of \vkg scenarios in order to understand how patterns occur in practice, and with which frequency. To this purpose, we have gathered 6 different scenarios\footnote{Available here: \url{https://github.com/ontop/ontop-examples/tree/master/dke-2022-mapping-patterns/}}, coming either from the literature on \vkgs, or from actual real-world applications. Table~\ref{t:scenarios} shows the results of our analysis, and for each cell pattern/scenario, it reports the number of applications of that pattern over that scenario (leftmost number in the cell) and the number of mappings involved (rightmost number in the cell). The last column in the table reports total numbers. We have manually classified a total of $1559$ mapping assertions, falling in $407$ applications of the described patterns. Of these applications, about $52.8\%$ are of schema-driven patterns, $44.7\%$ of data-driven patterns, and $2.5\%$ are of patterns falling outside of our categorization. In the remainder of this section we describe the detailed results for each scenario. In~\cite{DBLP:conf/caise/CalvaneseGHLM0S21} we present a similar evaluation, but restricted to schema-driven patterns and based on the results of an automated tool able to discover patterns starting from DB schemas.

\mypar{Berlin Sparql Benchmark (\bsbm)~\cite{BiSc09}}
This scenario is built around an e-commerce use case in which products are offered by vendors and consumers review them. Such benchmark does not natively come with mappings, but these have been created in different works belonging to the \vkg literature. We analyzed those in~\cite{SWJ-2019}. The ontology in \bsbm reflects quite precisely the actual organization of data in the DB. Due to this, each mapping falls into one of the patterns we identified. Notably, in the DB foreign key constraints are not specified. Therefore, we notice a number of applications of data-driven patterns, which cannot be captured by simple approaches based on \dm.

\mypar{NPD Benchmark (\npd)~\cite{LRXC15}}
This scenario is built around the domain of oil and gas extraction. It presents the highest number of mappings ($>$1k). The majority of these were automatically generated, and fall under \dm or schema-driven patterns. There are, however, numerous exceptions.
Mainly, there are a few denormalized tables which require the use of the \pat{DR1Nm} pattern, such as for the following mapping:

\begin{lstlisting}
target		npd:quadrant/{wlbNamePart1} a npdv:Quadrant .
source		SELECT "wlbNamePart1" FROM "wellbore_development_all"
\end{lstlisting}

A quadrant is not an entity in the DB schema (because \ex{wlbNamePart1} is not a key
of \ex{wellbore\_development\_all}), but it is represented as a class in the ontology. Moreover, quadrants have themselves their own data (resp., object) properties, triggering the application of other patterns in composition with \pat{DR1Nm}.

\mypar{University Ontology Benchmark (UOBM)~\cite{uobm}}
This scenario is built around the academic domain. Such benchmark provides a tool to automatically generate OWL ontologies, but does not include mappings nor a DB instance. These two have been manually crafted in~\cite{BCSS*16}, by reverse-engineering the ontology. The mappings in this setting are quite interesting, and are mostly data-driven as witnessed by the many applications of the clustering patterns. One critical aspect about these mappings is the use of a sophisticated version of the identifier alignment pattern modifier. Specifically, the table \ex{People} has the following primary key:
\begin{lstlisting}
PRIMARY KEY (ID,deptID,uniID,role) 
\end{lstlisting}
Table \ex{GraduateStudent}, which at the conceptual level corresponds to a subclass of the class \ex{People}, has the following key which is incompatible with the one of the superclass:
\begin{lstlisting}
PRIMARY KEY (studentID,deptID,uniID)
\end{lstlisting}
The subclass relation between \ex{People} and \ex{GraduateStudents} requires the two keys to be aligned. This is done ``artificially'', in the sense that the missing field \ex{role} is created on-the--fly by the mapping:
\begin{lstlisting}
target		<http://www.Dept{deptID}.Univ{univID}.edu/{role}{studID}> a :GraduateStudent . 
source		SELECT deptID, univID, studID, 'GraduateStudent' as role FROM GraduateStudents
\end{lstlisting}
\mypar{Suedtirol OpenData (\stod)\footnoteurl{https://github.com/dinglinfang/suedTirolOpenDataOBDA}} This is an application scenario coming from the turism domain. The ontology has been created independently from the DB. Moreover, the DB is itself highly de-normalized, since it is essentially a relational rendering of a JSON file. These aspects have a direct impact on the patterns we observed. In particular, we identified several applications of the \pat{DR11m} pattern, which, as we discussed, poses a huge challenge to automatic generation of mappings. Further complications arise from a number of applications of the value invention pattern modifier, which appears quite often in the form, for instance, of language tags:
\begin{lstlisting}
target		:mun/mun={istat_code} a :Municipality ; rdfs:label {name_i}@it, {name_d}@de . 
source		SELECT istat_code, name_i, name_d FROM municipalities
\end{lstlisting}
\mypar{Open Data Hub VKG (\odh)\footnoteurl{https://sparql.opendatahub.bz.it/}} This setting is the one behind the SPARQL endpoint located at the Open Data Hub portal from the Province of Bozen-Bolzano (Italy). This setting is also a denormalized one, and the same considerations we made for \stod apply to this setting as well.

\mypar{Cordis\footnote{\url{https://www.sirisacademic.com/wb/}}}
This setting is provided by SIRIS Academic S.L., a consultancy company specialized in higher education and research, and is designed around the domain of competitive research projects. As opposed to the previous two scenarios, this one comes with a well-structured relational schema, which reflects in a number of applications of schema patterns. Although in this scenario we have DB views, such views have explicit constraints defined on them (such as, \texttt{UNIQUE} constraints in SQL) that allow for the application of schema patterns.

\begin{table}[t!]
  \centering
{  \scriptsize
  \setlength{\tabcolsep}{5.5pt}
  \begin{tabular}{l|rrrrrrrrrrrrrr}
    \toprule
    & \multicolumn{2}{c}{\bsbm} & \multicolumn{2}{c}{\npd} & \multicolumn{2}{c}{\uobm} & \multicolumn{2}{c}{\odh} & \multicolumn{2}{c}{\stod} & \multicolumn{2}{c}{\cordis} & \multicolumn{2}{c}{Total} \\
    & & & & & & & & & & & & & & \\
    \midrule
    \pat{SE} & 8 & 52 & 34 & 406 & 8 & 16 & 10 & 43 & 8 & 37 & 13 & 60 & 81 & 614 \\
    \pat{SR} & \nn & \nn & \nn & \nn & 2 & 2 & \nn & \nn & \nn & \nn & 3 & 3 & 5 & 5 \\
    \pat{SRm} & 8 & 8 & 74 & 74 & 5 & 5 & \nn & \nn & 7 & 7 & 10 & 10 & 104 & 104 \\
    \pat{SEw/DEw} & \nn & \nn & 30 & 266 & 1 & 1 & \nn & \nn & \nn & \nn & \nn & \nn & 31 & 267 \\
    \pat{SRR} & \nn & \nn & 1 & 12 & \nn & \nn & \nn & \nn & \nn & \nn & 1 & 16 & 2 & 28 \\
    \pat{SH} & \nn & \nn & 3 & 132 & \nn & \nn & \nn & \nn & \nn & \nn & \nn & \nn & 3 & 132 \\
    \pat{DE} & \nn & \nn & \nn & \nn & \nn & \nn & \nn & \nn & 3 & 7 & 4 & 9 & 7 & 16 \\
    \pat{DRm} & 5 & 5 & 17 & 17 & 36 & 36 & 2 & 2 & 1 & 1 & 2 & 2 & 63 & 63 \\
    \pat{DH} & \nn & \nn & \nn & \nn & 5 & 9 & \nn & \nn & \nn & \nn & \nn & \nn & 5 & 9 \\
    \pat{DRR} & 2 & 2 & \nn & \nn & \nn & \nn & \nn & \nn & \nn & \nn & \nn & \nn & 2 & 2 \\
    \pat{DR1Nm} & 4 & 4 & 19 & 54 & \nn & \nn & 6 & 78 & 14 & 29 & 1 & 1 & 44 & 166 \\
    \pat{DH01} & \nn & \nn & \nn & \nn & \nn & \nn & \nn & \nn & \nn & \nn & 1 & 2 & 1 & 2 \\
    \pat{CE2C} & \nn & \nn & 11 & 82 & 6 & 19 & 5 & 23 & \nn & \nn & 1 & 12 & 23 & 136 \\
    \pat{CE2D} & \nn & \nn & 23 & 49 & \nn & \nn & \nn & \nn & \nn & \nn & \nn & \nn & 23 &49 \\
    \pat{CE2O} & \nn & \nn & 13 & 148 & \nn & \nn & \nn & \nn & \nn & \nn & \nn & \nn & 13 & 148 \\
    \pat{UNKNOWN} & \nn & \nn & 3 & 6 & 2 & 13 & 1 & 4 & 1 & 12 & 4 & 9 & 11 & 44 \\
    \bottomrule
  \end{tabular}
  }
  \caption{\label{t:scenarios} Occurrences of Mapping Patterns over the considered scenarios.}
\end{table}


%% file: 7-related-long.tex
\section{Related Work}
\label{sec:related}

In the last two decades a plethora of tools and approaches have been developed
to bootstrap an ontology and mappings from a DB. The approaches in the
literature differ in terms of the overall \emph{purposes} of the bootstrapping
(e.g., VKGs, data integration, ontology learning, check of DB schema
constraints using ontology reasoning), the \emph{ontology and mapping
 languages} in place (e.g., \owl profiles or RDFS, as ontology languages, and
R2RML or custom languages, for the specification of mappings), the different
focus on \emph{direct and/or complex mappings}, and the assumed \emph{level of
 automation}.  The majority of the most recent approaches closely follow \dm,
deriving ontologies that mirror the structure of the input DB, and are equipped with further ontology-to-ontology mapping techniques and custom mapping definition interfaces and languages in order to support the user in aligning the extracted ontology with domain-specific ontologies whose concepts and relationships pertain to a given domain of interest.

A notable example in this category is represented by the D2RQ system~\cite{bizer2004d2rq}. Once in its automated mode, D2RQ relies on the implementation of \dm, plus of additional reverse engineering methods for the discovery of many-to-many relationships and the translation of foreign keys into object properties. Manual and semi-automatic modes are also allowed in the system. In these cases, the user has available an RDF-compatible mapping language that can be used to map subset only of a relation, specify the mechanism for the generation of the individual URIs in the ontology and schemes for the translation of database values.
Another tool that complements the automatic extraction of database-to-ontology mapping is Ultrawrap Mapper~\cite{DBLP:conf/semweb/SequedaM15}. The main aim of this system is to hide the complexity of the R2RML language to the user by offering a bootstrapping of the mapping process which is based on an enhanced version of \dm. The user can interact with the system in several ways: by choosing which tables and attributes  of the database are to be mapped, refining the automatically extracted ontology, suggesting that specific data values have to be treated as ontology classes, specifying domain-driven SQL views to represent new concepts and, finally, by uploading a target domain-ontology for which specific mapping recommendations will be provided based on ontology matching techniques.

A special category of boostrapping tools that is worth to mention the one including those systems that implement extensions of \dm, while keeping a semi-automatic or fully automatic approach to mapping generation. A system that is representative of this category is BootOX~\cite{JKZH*15}, which relies on the R2RML language to produce direct mappings. BootOX implements a (fully-automatic) schema-driven bootstrapping. It aims at facilitating creation of an ontology by automatic extraction through bootstrapping technique from relational databases. Differently from other systems, BootOX covers all the \owl profiles of the output ontologies and suggest how to implement the ontology axioms resulting from a given mapping according to them (e.g., ``functionality'' or ``inverse functionality'' of a given data property are covered by the \owlrl and \owlel profiles only, as well as ``key axiom'' for a given class). The user can select the preferred \owl profile, as well as special characteristics of the extracted properties (e.g., symmetry, transitiveness or reflexivity) which may not be explicitly present in the  source database. Besides \dm, BootOX supports the user in building complex mappings based on selection and/or join operators. The proposed extensions to \dm are about mapping patterns which resolve into sub-class relationships and class hierarchies. To this aim, clusters of tuples in a table are detected by means of numerical vectors distance metrics, as well as subsets of the attributes of a table (especially when tables are not in BCNF normal form) with repeated values, and special evaluation of the application of outer join operations among tables. Still, the generation of these complex mappings heavily rely on the user intervention, e.g., in the naming of the newly discovered classes and properties.
The generation of provenance mappings (at different levels of granularity) is also an extension to \dm provided by BootOX: provenance information, such as the source database from which the information is extracted or the table and column identifiers, are modeled in the mapping assertions. BootOX is able to perform ad-hoc alignment with a target ontology using the LogMap system~\cite{solimando2014detecting,jimenez2012large,jimenez2011logmap}.

MIRROR~\cite{mirror} is also a tool for the automatic generation of R2RML mappings that extends \dm. Taking a DB as input (which is assumed to be, at least, in 3NF), MIRROR generates two groups of mappings: a first one strictly derived from the application of \dm, and a second one complementing the first with additional transformations that are extracted from information, such as class hierarchies and many-to-many relationships, that are usually not directly derivable from a database schema. From the point of view of the present contribution, the MIRROR categorization of the relationships between tables (two or more) that can be observed in the physical implementation of a relational database is especially interesting. Most of the categories identified in MIRROR can be linked with the mapping patterns introduced in the present paper, as we have seen in Section~\ref{sec:mappings}.

Rather than providing a bootstrappping algorithm, works~\cite{SequedaArenasDM} and~\cite{TG21} have instead studied the foundational framework underlying \dm and proved a number of results. The former presents a \emph{Direct Mapping} (different from \dm) that enjoys \emph{query preservation}, that is, every query that can be posed over the original DB schema can be expressed as a query over the mapped KG. It also shows that Direct Mapping is not \emph{lossless}, due to the impossibility of rendering constraints such as foreign and primary keys in the \owl language. Similar conclusions are reached by the latter work, that instead of \owl uses \emph{\shacl constraints}~\cite{W3Crec-SHACL}, in order to better translate DB (closed-world) constraints into their KG counterparts.

Bootstrapping approaches based on learning techniques and similarity measures are also present in the literature.
In~\cite{cerbah2014,cerbah2008mining,cerbah2008learning}, or instance, ontology learning techniques are used to mine the source data and discover patterns that suggest how to specify the target ontology. The core of RDBToOnto~\cite{cerbah2008learning} is based on the learning algorithm RTAXON~\cite{cerbah2008mining}, which performs hierarchy mining in the data to identify categorization patterns from which class hierarchies can be generated. RDBToOnto (which is currently part of the commercial portfolio of software by the~\href{http://redcoolmedia.net}{http://redcoolmedia.net} company) also allows users to iteratively refine the results provided by the automatic generation of the ontology, by suggesting a number of predefined constraints to be added (e.g., the selection of relevant categorization attributes and user-defined instance naming). The tool also includes a database normalisation step driven by the identification of specific inclusion dependencies by the user, whose main aim is to try to eliminate data duplication and, more in general, redundancy due to bad design in the source tables. Unfortunately, we have not been able to find a description of the mappings generated by the tool, and this prevents us from a deeper comparison with our approach.

In schema matching literature, simple rule-based mappings are used to create a uniform representation of the data sources to be matched, may they be schemata or ontologies~\cite{aumueller2005schema,gal2004ontobuilder,pinkel2013incmap}. For example, in COMA++, concept hierarchies, attributes and relationship types are mapped into generic model representation based on directed acyclic graphs. Using such mappings, schema matchers are applied to the uniform representation to create a matching result. Similarly, IncMap relies on a graph structure called IncGraph to represent schema elements from both ontologies and relational schemata in a unified way. Therefore, the main algorithmic task is to convert the ontology as well as the relational schema into the unified IncGraph representation. IncMap then computes ranked correspondences between elements of the two graphs using lexical and structural similarities, based on the Similarity Flooding algorithm, and converts the correspondences into direct mappings between the ontology and schema. The entire process is semi-automatic and each suggestion needs to be accepted by the user (human inputs are used to rerank correspondences after each round of interaction).

As written in~\cite{gupta2012karma}, Karma provides support for extracting data from a variety of sources (relational databases, CSV files, JSON, and XML), for cleaning and normalizing data, for mapping it to a target vocabulary, for integrating multiple data sources, and for publishing in a variety of formats (CSV, KML, and RDF).
Karma does not support a fully automatic mapping generation: it supports the users with a sophisticated user interface where a OWL-specified vocabulary, one of more data sources and a database of so-called semantic types are assumed as input. The algorithmic core of the system, which computes the relationships among the schema elements of a source, is based on conditional random fields (CRF)~\cite{CRFields} and a Steiner tree algorithm. The idea is that Karma automatically infers the semantic types (i.e. pairs made of a concept and/or a data property of the domain ontology, such as, <Person,name> or <Artwork,title>) that has been trained to identify in the source tables, whereas it asks the user to explicitly specify them whenever this process fails. In a second step, those semantic labelled source attributes are related each other in terms of the properties of the target ontology in order to reconstruct the overall semantic model of the data source (for instance, in the case above, <painter> is then suggested to be the right relationship between <Person> and <Artwork>, instead of <owner> or <sculptor>). More details about the way Karma exploits the knowledge from a domain ontology and the semantic models of previously modelled sources to automatically learn a rich semantic model for a new source can be found in~\cite{knoblock2012semi} and~\cite{taheriyan2016learning}. These papers focus on the characteristics of the algorithms that are responsible for the generation of the semantic models of the sources but the final mappings are never explicitly specified according to a standard mapping language. Nonetheless, the CRF-based approach is declared to encompass other existing approaches base on schema matching techniques that have been used to identify the semantic types of source attributes by comparing them with already labeled ones, such as~\cite{doan2000learning}, or~\cite{lerman2007semantic} which is based on the learning of regular expression-like rules for data in the source columns.

Although the work starts from a setting that is different from the one assumed in this paper, it is worth mentioning the work in~\cite{pequeno2014specifying}. The paper focuses on the automatic compiling of R2RML mappings once a set of algebraic correspondences have been manually specified by the user between a relational source and a target ontology. The so-called ``Correspondence Assertions'' are a set of algebraic assertions that are meant to express data-metadata mapping, mappings containing custom value functions (e.g., use transformation functions to change the data formats) and union, intersection and difference between tables. SQL views are then crated on the basis of the specified user assertions and used as an intermediate step to automatically compile the final R2RML mappings. The design and implementation of an authoring tool supporting the user in the specification of the algebraic correspondence assertions is still underway.

As for a systematic categorization of mappings, works~\cite{sequeda2012,sequeda-book} are very closely related ours, as they also introduce a catalog of mapping patterns. However, there are major differences between such work and the present, namely in these works:
\begin{compactitem}
\item patterns are not formalized, and presented in a ``by-example'' fashion following the R2RML syntax;
\item patterns are derived from ``commonly-occurring mapping problems'' based on the experience of the authors, whereas in this work patterns are derived from conceptual modeling and database design principles;
\item patterns are not evaluated against a number of different real-world, and complex scenarios over heterogeneous domains and design practices as it was done here.
\end{compactitem}

Other works providing a systematic categorization of mappings in \vkgs are \cite{mirror} and~\cite{JKZH*15}. However, the focus of such works is aimed at supporting bootstrapping of mappings. Other contributions restrict their attention to the algorithms behind the generation of mappings, notably~\cite{pinkel2013incmap,CCKK*17,DBLP:conf/semweb/SequedaM15} for the R2RML language.

The following are surveys and comparative analyses~\cite{sequeda2011survey,spanos2012bringing,Mogotlane-article,Pinkel2016RODIBR,Su-Cheng-review} where the interested reader can find the possibility to further explore on the tools and techniques mentioned here. We finally notice that, in our review, we did not find any study introducing an in depth analysis of existing real scenarios of DB-to-ontology mapping, as we do in the present paper, aimed at showing that the identified categories actually reflect the real design choices and methodologies in use by the mapping designers.


%% file: 8-conclusions.tex
\section{Conclusion and Future Work}
\label{sec:conclusions}

In this work we identified and formally specified a number of mapping patterns emerging when linking DBs to ontologies in a typical \vkg setting. Our patterns are grounded to well-established practices of DB design, and render explicit the connection between the conceptual model, the DB schema, and the ontology. We argue that the organization in patterns can enable a number of relevant tasks, apart from the classic one of bootstrapping mappings in an incomplete \vkg scenario. Through a systematic analysis of various \vkg scenarios, ranging from benchmarks to real world and denormalized ones, we observed that the patterns we formalized occur in practice, and capture most cases.

This work is only a first step, with respect to both categorization of patterns, and their actual use. Regarding the former, our plan is to extend this initial catalog with more advanced data-driven patterns, such as those deriving from other kinds of transformations for the hierarchies in the conceptual model (e.g., the case where all children and their attributes are fully merged in the parent entity, or the inverse case). to better explore the interaction between patterns and pattern modifiers, such as value invention or identifier alignment. Regarding the latter, in this paper we have used patterns to investigate, and highlight, the specific problems to address when setting-up a \vkg scenario. We plan to investigate solutions to these problems, by exploiting approaches from other fields, e.g., schema matching.

One point we did not formally investigate in this work is about what properties are guaranteed by applying our patterns (e.g., losslessness or query preservation). Since our patterns are grounded in textbook methodologies from DB design, such as standard translations of ER-diagrams into the relational model, we believe that query preservation might work without any substantial modification. We plan to investigate these aspects, in the same spirit as the literature in~\cite{SequedaArenasDM} or~\cite{TG21}, in a future work.